\documentclass[final,3p,times,twocolumn]{elsarticle}
\usepackage{amssymb}
\usepackage{amsmath}
\usepackage{hyperref}
\usepackage{color}
\usepackage{pifont}
\usepackage{booktabs}
\usepackage{multirow}
\usepackage[normalem]{ulem}
\useunder{\uline}{\ul}{}
\usepackage{bbding}
\usepackage{algorithm}
\usepackage{algorithmic}
\journal{ISPRS Journal of Photogrammetry and Remote Sensing}

\newtheorem{definition}{Definition}
\newtheorem{assumption}{Assumption}

\begin{document}

\begin{frontmatter}

\title{Homogeneous Tokenizer Matters: Homogeneous Visual Tokenizer for Remote Sensing Image Understanding}

\author[csu, XJL]{Run Shao}
\ead{shaorun@csu.edu.cn}

\author[csu]{Zhaoyang Zhang}
\author[csu]{Chao Tao}
\author[csu]{Yunsheng Zhang}
\author[csu]{Chengli Peng}

\author[csu, XJL]{Haifeng Li\corref{cor1}}
\ead{lihaifeng@csu.edu.cn}

\affiliation[csu]{organization={School of Geosciences and Info-Physics, Central South University, No. 932 South Lushan Road, Changsha, 410083, Hunan, China}}

\affiliation[XJL]{organization={Xiangjiang Laboratory, No. 569, YueLu Avenue, Changsha, 410083, Hunan, China}}

\cortext[cor1]{Corresponding author}

\begin{abstract}
On the basis of the transformer architecture and the pretext task of "next-token prediction", multimodal large language models (MLLMs) are revolutionizing the paradigm in the field of remote sensing image understanding. However, the tokenizer, as one of the fundamental components of MLLMs, has long been overlooked or even misunderstood in visual tasks. A key factor contributing to the great comprehension power of large language models is that natural language tokenizers utilize meaningful words or subwords as the basic elements of language. In contrast, mainstream visual tokenizers, represented by patch-based methods such as Patch Embed, rely on meaningless rectangular patches as basic elements of vision. Analogous to words or subwords in language, we define semantically independent regions (SIRs) for vision and then propose two properties that an ideal visual tokenizer should possess: (1) homogeneity, where SIRs serve as the basic elements of vision, and (2) adaptivity, which allows for a flexible number of tokens to accommodate images of any size and tasks of any granularity. On this basis, we design a simple HOmogeneous visual tOKenizer: HOOK. HOOK consists of two modules: an object perception module (OPM) and an object vectorization module (OVM). To achieve homogeneity, the OPM splits the image into 4$\times$4 pixel seeds and then uses a self-attention mechanism to identify SIRs. The OVM employs cross-attention to merge seeds within the same SIR. To achieve adaptability, the OVM predefines a variable number of learnable vectors as cross-attention queries, allowing for the adjustment of the token quantity. We conducted experiments on the NWPU-RESISC45, WHU-RS19, and NaSC-TG2 classification datasets for sparse tasks and the GID5 and DGLCC segmentation datasets for dense tasks. The results show that the visual tokens obtained by HOOK correspond to individual objects, thereby verifying their homogeneity. Compared with randomly initialized or pretrained Patch Embed, which required more than one hundred tokens per image, HOOK required only 6 and 8 tokens for sparse and dense tasks, respectively, resulting in performance improvements of 2\% to 10\% and efficiency improvements of 1.5 to 2.8 times. The homogeneity and adaptability of the proposed approach provide new perspectives for the study of visual tokenizers. Guided by these principles, the developed HOOK has the potential to replace traditional Patch Embed. The code is available at https://github.com/GeoX-Lab/Hook.
\end{abstract}

\begin{keyword}

Remote Sensing Image Understanding \sep Visual Tokenizer \sep Homogeneous \sep Semantically Independent Region \sep Visual Transformer Model

Citation: Run Shao, Zhaoyang Zhang, Chao Tao, Yunsheng Zhang, Chengli Peng, Haifeng Li. Homogeneous Tokenizer Matters: Homogeneous Visual Tokenizer for Remote Sensing Image Understanding. ISPRS Journal of Photogrammetry and Remote Sensing. 2024, 18: 294-310. DOI: 10.1016/j.isprsjprs.2024.09.009.
\end{keyword}

\end{frontmatter}

\begin{figure*}[!t]
	  \centering
			\includegraphics[width=1.0\textwidth]{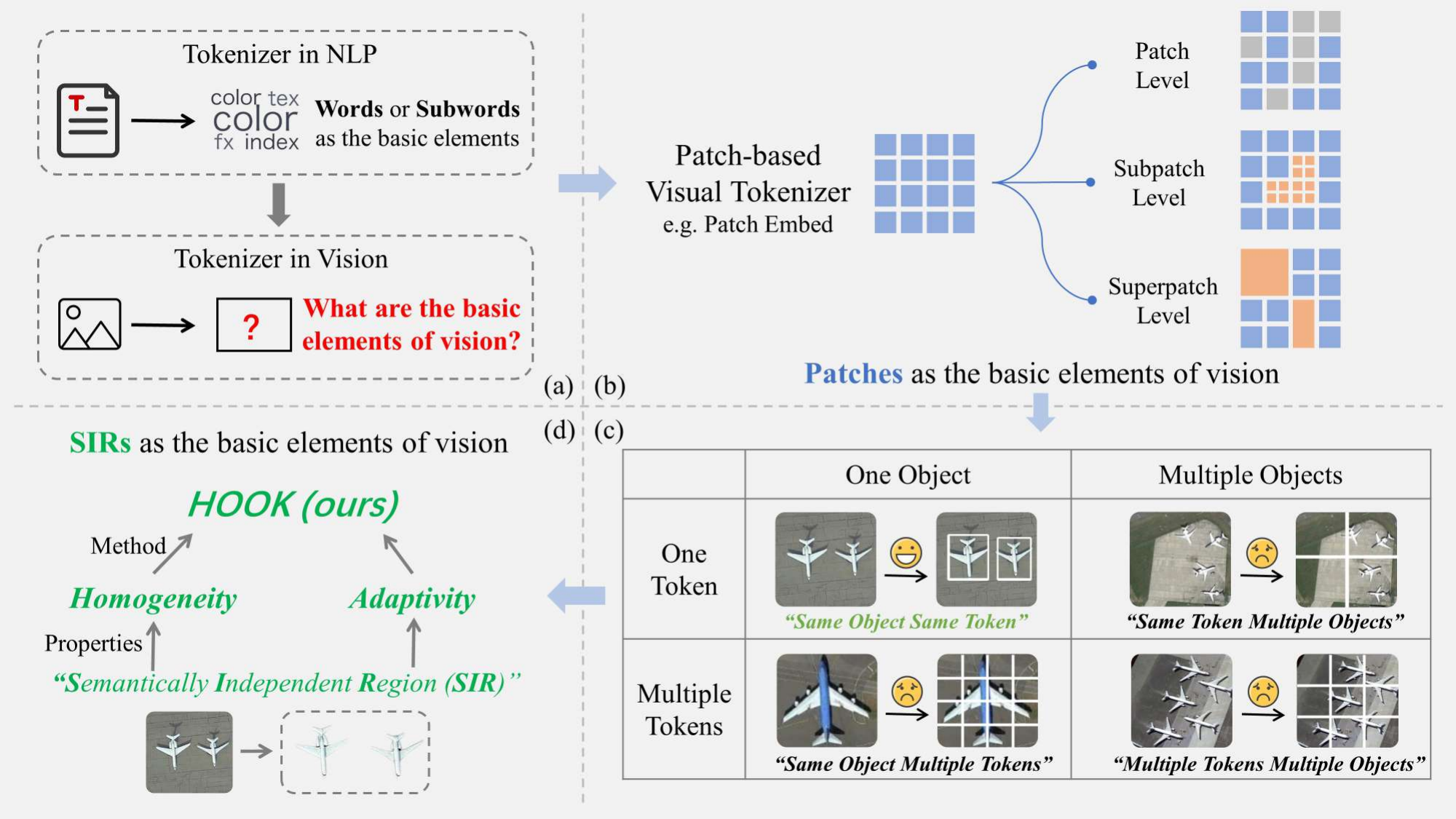}
        \caption{(a) A natural language tokenizer uses words or subwords as the basic elements of language. Similarly, for visual tokenizers, this work attempts to answer a fundamental question: What are the basic elements of vision? (b) The current mainstream visual tokenizers are patch-based methods, represented by Patch Embed, which can be categorized into three types on the basis of hierarchy: patch level, subpatch level, and superpatch level. Their commonality lies in the use of {\color{red}{\textbf{patches as the basic elements of vision}}}. (c) There exists a confusion matrix for tokens and objects. The "same object multiple tokens" leads to incomplete object features, whereas the "same token multiple objects" leads to unclear relationships between objects. The "multiple tokens multiple objects" inherits the drawbacks of the above two approaches. Patch-based methods inherently struggle to achieve the ideal "same token same object" scenario. (d) We define the concept of semantically independent region (SIRs) and propose two properties of an ideal visual tokenizer: homogeneity and adaptability. We design a simple \textbf{HO}mogeneous visual t\textbf{OK}enizer, HOOK, where {\color{red}{\textbf{SIRs as the basic elements of vision}}}.}
\label{fig:overview}
\end{figure*}

\section{Introduction}
\label{introdcution}
Large language models, such as the GPT series\cite{radford2018improving, brown2020language, radford2019language, achiam2023gpt}, Llama\cite{touvron2023llama, touvron2023llama2}, and PaLM\cite{chowdhery2023palm}, which are built on the transformer architecture and trained with the "next-token prediction" objective, have demonstrated powerful natural language understanding, reasoning, and generalization capabilities. Subsequently, the development of multimodal large language models\cite{liu2024visual, bai2023qwen, zhan2024skyeyegpt, guo2024remote, zhang2024earthgpt} has dramatically reshaped the research landscape in various fields, including remote sensing image understanding.

Tokenizers are a foundational and essential component of multimodal large language models. They are designed to identify basic elements of input data and convert them into a sequence of tokens. Natural language tokenizers (NL tokenizers) are fundamental for machine understanding of language and have been extensively researched in the field of natural language processing\cite{mikolov2013efficient, pennington2014glove, kudo2018sentencepiece, kudo2018subword}. As shown in Figure \ref{fig:overview}-a, one key factor contributing to the great comprehension power of large language models is that natural language tokenizers utilize meaningful words or subwords as the basic elements of language. However, visual tokenizers, which play a similar role in visual foundation models and multimodal large models, have long been overlooked, underestimated, and even misunderstood.

A visual tokenizer is aimed at identifying the basic elements of an image and tokenizing them into a sequence of tokens. \textit{What are the basic elements of vision?} As shown in Figure \ref{fig:overview}-b, patch-based methods, represented by Patch Embed\cite{dosovitskiy2020image}, are the most popular implementations of visual tokenizers. The fundamental characteristic of these methods is the use of rectangular \textbf{patches as the basic elements of vision}. The advantage of patch-based methods lies in their simplicity and efficiency, but they have two prominent issues for remote sensing images:

(1) Rectangular patches fail to match irregular and complex objects. As shown in Figure \ref{fig:overview}-c, according to the confusion matrix for tokens and objects, if multiple objects are aggregated within a single token, the model may struggle to learn the relationships between the objects, whereas if a single object is dispersed across multiple tokens, the model may struggle to learn the complete features of the individual object. We refer to these two scenarios as the \textit{"same token multiple objects"} and \textit{"same object multiple tokens"} scenarios, respectively. Additionally, if multiple objects correspond to multiple tokens, i.e., \textit{"multiple tokens multiple objects"}, this scenario inherits the drawbacks of the previous scenarios. Ideally, the relationship between tokens and objects should be \textit{"same object same token"}, but owing to the limitations of rectangular patches, patch-based methods fundamentally cannot achieve this ideal scenario.

(2) Efficiency issues arising from the fixed and redundant number of tokens limit the application of transformer-based models to remote sensing images. For example, when the patch size is 16 $\times$ 16, a 224 $\times$ 224 remote sensing image corresponds to 196 tokens, and the number of tokens grows quadratically with increasing image size. Furthermore, in transformer models, the computational complexity of the attention mechanism is $O(N^2)$, which indicates that the computational cost of the attention mechanism also scales quadratically with the number of tokens. This feature makes it challenging for transformer-based remote sensing models to handle high-resolution remote sensing images.

Therefore, meaningless rectangular patches are not suitable as the basic elements of images. Analogous to meaningful words or subwords that serve as the basic elements of natural language, an intuitive idea is to identify certain semantically meaningful regions in the image as the basic elements. Therefore, as illustrated in Figure \ref{fig:overview}-d, we define semantically independent regions (SIRs) and propose two properties that an ideal visual tokenizer should possess:

(1) \textbf{Homogeneity}: SIRs serve as the basic elements of vision. In this context, an image is a union of all SIRs rather than meaningless rectangular patches.

(2) \textbf{Adaptivity}: The number of tokens should be flexibly adjusted to accommodate images of any size and tasks of varying granularity. Larger images and more fine-grained tasks require more visual tokens, whereas smaller images and simpler tasks need fewer visual tokens.

To develop an ideal visual tokenizer that meets the above properties, we define a confusion matrix for tokens and objects under strict definitions and constraints. We observe that the transformations among \textit{"same object same token"}, \textit{"same object multiple tokens"}, \textit{"same token multiple objects"}, and \textit{"multiple tokens multiple objects"} actually involve only two meta-operations: "split" and "merge". The process of obtaining homogeneous visual tokens can be abstracted as a routing selection problem, where two distinct general routes are "splitting and merging" and "merging and splitting".

On the basis of the "splitting and merging" route, we design a simple \textbf{HO}mogeneous visual t\textbf{OK}enizer: \textbf{HOOK}. HOOK consists of two modules: an object perception module (OPM) and an object vectorization module (OVM).

To achieve homogeneity, HOOK initially splits the image into the finest granularity possible and then gradually merges the split components to form SIRs. In the first step, the OPM uses convolutional blocks to generate 4 $\times$ 4-pixel seeds. The self-attention layers subsequently associate these seeds if they belong to the same SIR. Finally, the OVM employs cross-attention to merge these associated seeds into homogeneous visual tokens.

To achieve adaptivity, the OVM defines $N$ learnable vectors as queries for the cross-attention module, with the seeds output by the OPM serving as the keys and values. Here, $N$ is treated as a variable hyperparameter to allow for the arbitrary adjustment of the token quantity.

We tested the performance of HOOK on a sparse task via the NWPU-RESISC45\cite{cheng2017remote}, WHU-RS19\cite{Xia2010WHURS19, Dai2011WHURS19}, and NaSC-TG2\cite{NaSCTG2} classification datasets and on a dense task via the GID5\cite{Tong2020GID} and DGLCC\cite{DGLCC} semantic segmentation datasets. The experimental results show that the visual tokens obtained by HOOK correspond to individual objects, thereby verifying their homogeneity. Compared with randomly initialized or pretrained Patch Embed, HOOK achieved performance improvements ranging from 2\% to 10\%. Additionally, HOOK outperformed the baseline methods that we used for comparison, reaching state-of-the-art performance. In terms of efficiency, HOOK required only 6 tokens for the sparse task and 8 tokens for the dense task, whereas Patch Embed required hundreds of tokens for a single image. HOOK realized an overall efficiency improvement of 1.5 to 2.8 times.

Our contributions are summarized as follows:

(1) Our analyses and experiments show that the importance of visual tokenizers is far underestimated. Starting from the essence of tokenizers, we propose two fundamental properties that an ideal visual tokenizer should possess: homogeneity and adaptivity.

(2) We define a confusion matrix for tokens and objects to systematically analyse their intricate relationships. This analysis reveals that the process of constructing homogeneous tokens can be abstractly viewed as a routing selection problem involving two principal strategies: "splitting and merging" and "merging and splitting". This discovery provides new perspectives and ideas for research on visual tokenizers.

(3) Based on the "splitting and merging" strategy, we design a simple homogeneous visual tokenizer, HOOK, which offers an object-oriented, plug-and-play implementation for visual tokenizers. HOOK can obtain visual tokens from semantically independent regions in an image and is adapted to complex objects of any shape rather than just regular rectangular patches.

(4) The experimental results show that HOOK achieves both homogeneity and adaptivity. HOOK outperforms existing baselines in terms of performance and efficiency and has the potential to supplant patch-based methods as a new standard visual tokenizer.

The remainder of this paper is organized as follows: In Section \ref{related_work}, we review the advancements in visual tokenizers, which can be classified into two types, namely, patch-based and object-oriented, based on the different basic elements of vision. In Section \ref{theoretical_analysis}, we conduct a theoretical analysis of visual tokens. Section \ref{method} and Section \ref{experiments} introduce HOOK, a homogeneous visual tokenizer designed on the basis of the "splitting and merging" strategy, and discuss its performance via experiments. Section \ref{discussion} primarily delves into the significance of visual tokenizers, rethinks our HOOK, and outlines future research directions.

\section{Related Work}
\label{related_work}

\subsection{Patch-Based Visual Tokenizers}
\label{related_work_patch_based}

Patch-based methods use rectangular patches as the basic elements of images. Patch Embed, which is widely applied in patch-based methods, splits an image into nonoverlapping patches of a fixed size, typically 16$\times$16 pixels. The advantage of Patch Embed lies in its simplicity and efficiency. However, the use of redundant tokens and a fixed patch size have been identified in various works\cite{chen2021empirical, qian2022makes, ryoo2021tokenlearner, xiao2021early} as factors resulting in the instability of training in ViT models.

As shown in Figure \ref{fig:overview}-b, we categorize improved patch-based methods into three types on the basis of hierarchy: patch-level, subpatch-level, and superpatch-level methods.

\textbf{A. Patch-level Methods}

Patch-level methods are based on the argument that the visual tokens obtained through Patch Embed are highly redundant, and in practice, a model needs to select only a small portion of these tokens to recognize an image.

H. Yin et al. proposed A-ViT\cite{yin2021adavit}, where the first dimension of each token is defined as the probability of discarding the token, which enables the model to autonomously select tokens that aid in image recognition. Y. Tang et al. calculated the impact of each token at each layer on the final model output, identifying and removing redundant tokens\cite{tang2022patch}. Y. Rao et al. introduced DynamicViT\cite{rao2021dynamicvit}, and B. Pan et al. proposed IA-RED2\cite{pan2021ia}, both of which insert trainable prediction modules in the middle layers of the model for dynamic token selection. D. Marin et al. utilized pooling strategies to discard redundant tokens\cite{marin2023token}. Y. Liu et al. introduced PatchDropout\cite{liu2023patchdropout}, a simple method that randomly drops a certain percentage of patches during model training, leading to efficiency improvements without compromising test accuracy. Y. Liang et al. proposed a token reorganization module for identifying and merging tokens with minimal relevance to the CLS token, thereby reducing the number of tokens\cite{liang2022not}. Similarly, T. Wang et al. proposed PnP-DETR\cite{wang2021pnp}, which initially extracts features from the original image via ResNet as tokens and then identifies and merges background tokens, effectively enhancing model efficiency and object-level perception capabilities.

\textbf{B. Subpatch-level methods}

Subpatch-level methods are based on the argument that the implicit assumption in Patch Embed, namely, that all patches in an image should be treated equally, is unreasonable. In these methods, it is assumed that the patches corresponding to foreground regions in the image should be refined, meaning that more patches should be used to describe important areas of the image.

W. Chen et al. utilized class activation maps (CAMs) to identify salient regions in images and proposed the use of smaller patches for these significant areas to increase the overall number of patches\cite{chen2022authenticity}. Similarly, T. Ronen et al. employed GradCAM to locate salient regions and guide the generation of fine-grained patches via quadtree structures\cite{ronen2023vision}. M. Chen et al. introduced CF-ViT\cite{chen2023cf} and used confidence levels from a coarse-grained model to determine whether patches should undergo further processing. Y. Wang et al. also leveraged results from a coarse-grained model, continually densifying patches across the image until high-confidence recognition outcomes were obtained\cite{wang2021not}. X. Yue et al. proposed PS-ViT\cite{yue2021vision}, which employs progressive sampling to shift token positions towards the main regions of interest in the image, enabling a more detailed description of foreground elements through additional tokens.

\textbf{C. Superpatch-level methods}

Superpatch-level methods are based on the argument that the use of fixed-size patches restricts the model to capturing information at a single scale. These methods construct multiscale patches to increase the model's robustness across different scales.

C.-F. Chen et al. introduced CrossViT\cite{chen2021crossvit}, which uses two patches of different sizes to split an image, enabling the extraction of multiscale features from the image. L. Beyer et al. proposed FlexiViT\cite{beyer2023flexivit}, which enhances the model's robustness to different patch sizes by randomly interpolating parameters in the Patch Embed layer. S. H. Lee et al. presented shifted patch tokenization, which applies multiangle shifting and aggregation of patches to address the lack of local inductive bias in Patch Embed\cite{lee2021vision}.

\subsection{Object-Oriented Visual Tokenizers}
\label{related_work_object_oriented}

While the above methods have made significant progress in improving Patch Embed, patch-based approaches still cannot fundamentally overcome the limitations of rectangular patches. Researchers have begun to explore moving away from patches and instead reconstructing visual tokenizers on the basis of the objects in the images.

B. Wu et al. proposed VT\cite{wu2020visual}, which employs a simple convolution operation to assign each pixel in an image to one of several semantic groups, mapping them to visual tokens. However, VT lacks global semantic awareness of the image, potentially leading to an overreliance on local pixel characteristics such as colour and texture. T. Yang et al. introduced visual concept tokenization (VCT)\cite{yang2022visual}, which uses concept token reconstruction in a cleverly designed pretraining task to enable individual control over specific features by each visual token on simulated images. However, VCT requires pretraining and features a complex model structure. S. Qian et al. developed MoTo\cite{qian2022makes}, which perceives the semantic consistency association between tokens but does not represent them as homogeneous visual tokens. J. Mei et al. proposed SPFormer\cite{mei2024spformer}, which divides images into irregular, semantically homogenous regions. However, the method relies on superpixel algorithms and lacks plug-and-play functionality.

To address the limitations of the aforementioned approaches fundamentally, we revisit tokenization in natural language processing and propose two fundamental properties that a visual tokenizer should possess: homogeneity and adaptability. Building upon this premise, we introduce the HOmogeneous visual tOKenizer (HOOK), which has distinct advantages in terms of both performance and efficiency over baselines, including Patch Embed. HOOK has the potential to replace patch-based methods as a new foundational visual tokenizer.

\section{Theoretical Analysis}
\label{theoretical_analysis}

In this section, we conduct a theoretical analysis of visual tokens. In Section \ref{theoretical_analysis_why}, we emphasize the importance and necessity of homogeneity in visual tokenizers by examining the essence of tokenization. In Section \ref{theoretical_analysis_confusionMatrix}, we define a confusion matrix for tokens and objects under strict definitions and constraints. In Section \ref{theoretical_analysis_general_routing}, we propose two general routes for constructing homogeneous visual tokens on the basis of the confusion matrix that was previously established.

\subsection{Why do we need a homogeneous visual tokenizer?}
\label{theoretical_analysis_why}

The concept of a tokenizer originates from the field of natural language processing (NLP). Machines are unable to understand unstructured data, and the first step in enabling machines to comprehend language is to structurize unstructured language documents. NL tokenizers are among the most popular methods for achieving this process by breaking down language into smaller basic elements known as tokens.

\textbf{A. NL tokenizers}

In an NL tokenizer, the definition of a token is not fixed. Following the order from larger to smaller tokens, natural language tokenizers can be classified into sentence tokenizers, word tokenizers, subword tokenizers, and character tokenizers, among others. The size of the tokens involves a balance between efficiency and generalization:

(1) \textbf{Larger tokens are more efficient but have poorer generalizability.} For example, in a sentence tokenizer, each token represents a complete sentence, which allows for the efficient representation of a document with a small number of tokens. However, the corresponding vocabulary would need to contain all possible sentences, which is not feasible. As a result, sentence tokenizers struggle to handle out-of-vocabulary (OOV) issues and cannot generalize to unseen sentences.

(2) \textbf{Smaller tokens have stronger generalizability but are less efficient.} For example, in a character tokenizer, the vocabulary consists of only 26 letters, with necessary symbols. Any new word can be composed of these characters. In theory, a character tokenizer can be generalized to any text. However, because individual letters lack semantic meaning, machines struggle to learn words, sentences, and higher-level semantic information. Moreover, several tokens may be required to represent just one word.

Therefore, NL tokenizers need to define suitable basic elements of language to balance the conflicting goals of efficiency and generalization. The most popular NL tokenizers, such as Word2Vec\cite{mikolov2013efficient}, Byte-Pair Encoding (BPE)\cite{sennrich2015neural}, WordPiece\cite{schuster2012japanese}, Unigram\cite{kudo2018subword}, and others\cite{pennington2014glove, kudo2018sentencepiece}, belong to the category of word tokenizers or subword tokenizers, which use words or subwords as the basic elements of language. Words and subwords represent the smallest semantically independent elements in language, effectively striking a balance between efficiency and generalization.

\textbf{B. Visual tokenizers}

Research on visual tokenization has been driven by transformer-based visual models, which are aimed at identifying the basic elements of an image and splitting the image into a token sequence to meet the input requirements of transformer models. Like NL tokenizers, visual tokenizers also face the challenge of balancing efficiency and generalizability. We consider two extreme scenarios:

(1) Tokens that represent individual pixels. This approach offers the best generalizability because it can be generalized to any image within the same colour space. However, this method has the lowest efficiency: it requires 50,176 tokens to represent just one 224 $\times$ 224 image.

(2) Tokens that represent the entire image. Conversely, this approach maximizes efficiency but compromises generalizability.

In balancing these aspects, visual tokenizers are aimed at finding a suitable granularity of tokens that balances efficiency and generalizability for effectively processing images within transformer-based models.

An intuitive approach is to find an intermediate value between the two extreme scenarios mentioned above, similar to words and subwords in NL tokenizers. Notably, the most popular patch-based methods align with this approach. For example, Patch Embed uses patches of size 16 $\times$ 16 as an empirical middle ground to balance efficiency and generalizability. This is a fundamental reason why patch-based methods are prevalent in visual tokenization.

However, patch-based methods do not consider the differences between images and language. In language, words are inherently semantically independent basic elements, and subwords constructed through various optimization methods often consist of meaningful morphemes. However, in images, a fixed-size rectangular patch typically does not possess independent semantic meaning in most cases.

\textbf{C. Semantically Independent Regions}

As shown in Figure \ref{fig:overview}-d, on the basis of the observations and theoretical analysis discussed above, we define the semantically independent region (SIR) as a replacement for the fixed-size rectangular patch as the basic element of an image.

\begin{definition}
    Semantically independent region (SIR) refers to a distinct and self-contained region within an image that has a specific semantic meaning and is distinguishable from its surroundings.
\end{definition}

An SIR typically includes a single visual object, such as an airplane, a car, or a building, and it visually stands out from the surrounding environment and semantically represents a complete conceptual entity. Once semantically independent regions are defined, they are internally homogeneous in terms of semantics compared with interregions. This is why we believe that an ideal visual tokenizer should possess homogeneity.

In addition, in the definition of SIRs, "independence" is a dynamic concept. For example, an entire aircraft is an independent region in relation to the ground, wings are independent regions in relation to the aircraft, and flaps are independent regions in relation to the wings. Therefore, when addressing a specific remote sensing image, the image size and granularity of the task determine the \textbf{ideal} granularity of independence, whereas the number of tokens determines the \textbf{actual} granularity of independence. Additionally, an ideal visual tokenizer should be image-agnostic and task-agnostic. Therefore, the number of tokens should be adjusted to accommodate images of any size and tasks of any granularity. We refer to this property as the adaptivity of the visual tokenizer.

In conclusion, an ideal visual tokenizer should possess homogeneity (SIRs as the basic elements of vision) and adaptability (allowing for arbitrary adjustment of the token quantity).

\subsection{Confusion matrix for tokens and objects}
\label{theoretical_analysis_confusionMatrix}

\begin{figure*}[!t]
	  \centering
			\includegraphics[width=1.0\textwidth]{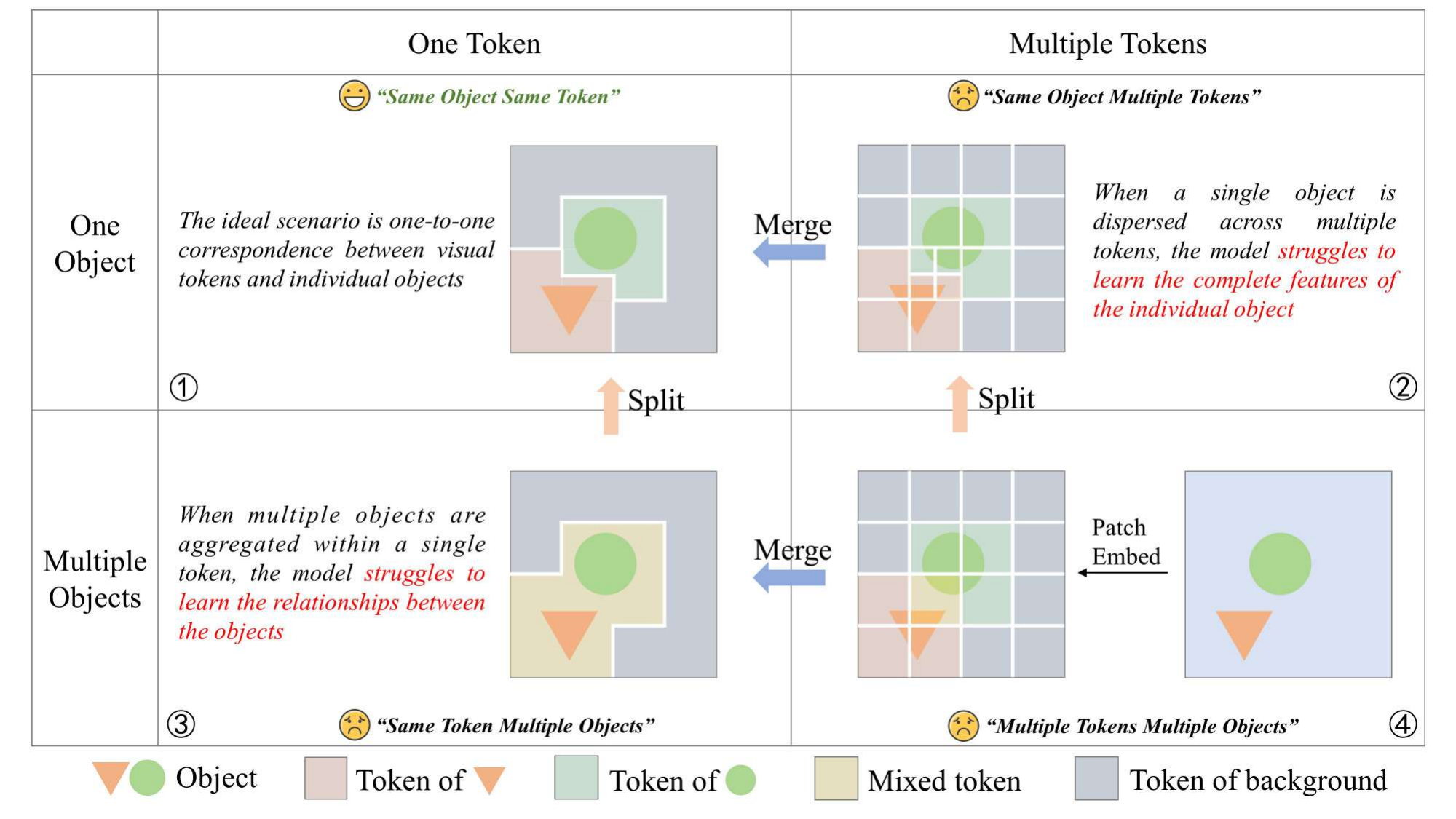}
        \caption{Confusion matrix of tokens and objects. In a simplified scenario and under strict definitions, the confusion matrix of tokens and objects reveals two general routes for constructing homogeneous visual tokens: (1) splitting and merging and (2) merging and splitting.}
\label{fig:confusionMatrix}
\end{figure*}

Figure \ref{fig:overview}-c illustrates the confusion matrix between tokens and objects in four real remote sensing images. To analyse the relationships among the four scenarios in the confusion matrix more clearly, we abstract and simplify the real scenes.

As shown in Figure \ref{fig:confusionMatrix}, we assume that each remote sensing image contains only two objects and that we are not concerned with the background. The tokens corresponding to the background regions are referred to as "background tokens". In fact, real remote sensing images containing multiple complex objects can be extrapolated from the simplified scenarios described above.

For rigour of presentation, we define the following concept:

\begin{definition}
    "Region of the token" refers to the pixel region on the original image that contains the semantic information encapsulated by a specific token.
\end{definition}

For ease of presentation and accuracy, we further define two types of relationships between the objects and regions of the tokens:

\begin{definition}
    "Cover" refers to the complete inclusion of an object in the region of the token on the image.
\end{definition}

\begin{definition}
    "Overlap" refers to the presence of an intersection between the region of the token and an object.
\end{definition}

From the above definitions, if the region of the token covers an object, they must overlap, whereas the reverse is \textbf{not} true.

On the basis of the above definitions, we can provide precise definitions for the four scenarios mentioned in Figure \ref{fig:overview}-c and Figure \ref{fig:confusionMatrix}:

\begin{definition}
    "Same object same token" means that, excluding background tokens, any region of the token covers exactly one object, and each region of the token is distinct from other regions.
\end{definition}

\begin{definition}
    "Same object multiple tokens" means that no region of a token covers an object, and at most, each region overlaps with one object.
\end{definition}

\begin{definition}
    "Same token multiple objects" means that, excluding background tokens, each region of a token covers at least two objects.
\end{definition}

\begin{definition}
    "Multiple tokens multiple objects" means that there is a region of each token that overlaps with two objects but covers at most one object.
\end{definition}

As shown in Figure \ref{fig:confusionMatrix}, the above definitions completely describe the relationships between tokens and objects.

\subsection{General route to homogeneous visual tokens}
\label{theoretical_analysis_general_routing}

In Section \ref{introdcution}, we show that \textit{"same object multiple tokens"} makes it difficult for the model to learn complete object features, that \textit{"same token multiple objects"} makes it difficult for the model to learn the relationships between objects, and that \textit{"multiple tokens multiple objects"} inherits the drawbacks of the above two cases. The ideal scenario between tokens and objects should be \textit{"same object same token"}, and we refer to this token as a homogeneous visual token. In other words, in the four squares of Figure \ref{fig:confusionMatrix}, only square 1 is ideal. Because the default visual tokenizer, Patch Embed, corresponds to square 4, \textbf{constructing the ideal visual token can be abstracted as a route selection problem from square 4 to square 1.}

According to the confusion matrix in Figure \ref{fig:confusionMatrix}, without considering diagonal routes, each step has only two choices: "move left" and "move up." "Move left" refers to keeping the same number of objects but reducing the number of tokens from multiple to one. For example, going from \ding{173} to \ding{172} signifies that multiple tokens corresponding to one object are reduced to one token corresponding to one object, with the basic operation for this step being to "merge" tokens. "Move up" signifies keeping the same number of tokens but reducing the number of objects from multiple to one. For example, going from \ding{174} to \ding{172} signifies that one token corresponding to multiple objects is reduced to one token corresponding to one object, with the basic operation for this step being to "split" tokens.

Thus, we obtain two general routes for constructing homogeneous visual tokens:

(1) Splitting and merging, which corresponds to the route \ding{175}
$\rightarrow$ \ding{173} $\rightarrow$ \ding{172} in Figure \ref{fig:confusionMatrix}: First, the tokens are split until no region of a token overlaps with two or more objects, achieving \textit{"same object multiple tokens"}. Then, the tokens are merged until no object corresponds to multiple tokens, achieving \textit{"same object same token"}.

(2) Merging and splitting, which corresponds to the route \ding{175} $\rightarrow $\ding{174} $\rightarrow$ \ding{172} in Figure \ref{fig:confusionMatrix}: First, the tokens are merged until no object corresponds to multiple tokens, achieving \textit{"same token multiple objects"}. Second, the tokens are split until no token corresponds to multiple objects, achieving \textit{"same object same token"}.

\section{Method}
\label{method}

Section \ref{theoretical_analysis_general_routing} outlines two general routes for constructing homogeneous visual tokens. On the basis of the "splitting and merging" route, we design a simple HOmogeneous visual tOKenizer: HOOK. In Section \ref{method_overview}, we provide an overview of the structure of HOOK. Section \ref{method_OPM} and Section \ref{method_OVM} detail two important modules in HOOK: the object perception module and the object vectorization module. In Section \ref{method_task}, we discuss how HOOK adapts to images of any size and tasks of any granularity.

\subsection{Overview}
\label{method_overview}

\begin{figure*}[!t]
	  \centering
	\includegraphics[width=1.0\textwidth]{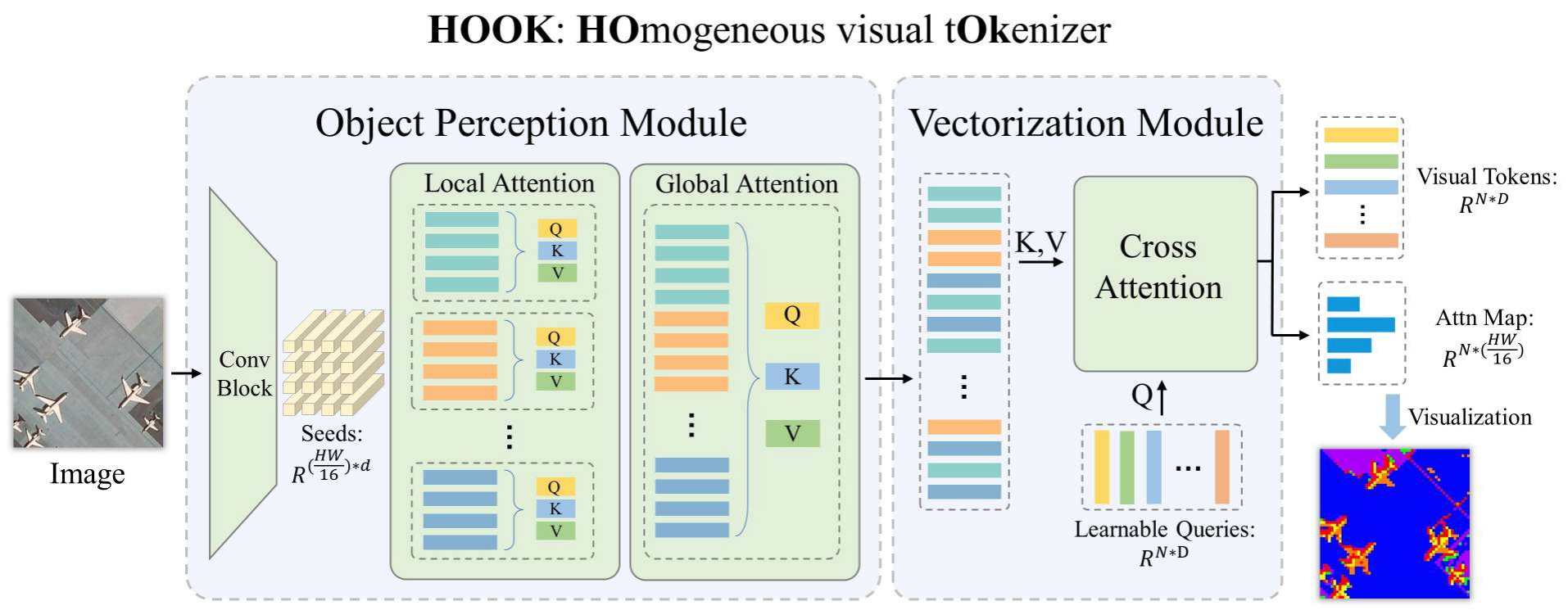}
        \caption{The architecture of HOOK consists of two modules: the object perception module (OPM), which is responsible for perceiving semantically independent regions, and the object vectorization module (OVM), which is responsible for vectorizing semantically independent regions into tokens.}
\label{fig:HOOK_overview}
\end{figure*}

The overall architecture of HOOK is shown in Figure \ref{fig:HOOK_overview}. HOOK can be viewed as a function $T$ that maps an image $I$ to $N$ D-dimensional vectors, with each vector representing a token, as follows:
\begin{equation}
    t = T(I)
\end{equation}
where $t\in \mathbb{R}^{N \times D}$ and $I\in \mathbb{R}^{H \times W \times C}$.

HOOK consists of two modules: the object perception module (OPM) and the object vectorization module (OVM).

The OPM is aimed at perceiving semantically independent regions within the image. Specifically, the OPM initially splits the image into several 4 $\times$ 4 pixel-sized seeds via convolutional blocks, which are followed by stacked local and global self-attention layers to expand the seeds into semantically independent regions (as detailed in Section \ref{method_OPM}). The process above can be formalized as follows:
\begin{equation}
    f = P(I)
\end{equation}
where $P$ represents the object perception module and where $f\in \mathbb{R}^{\frac{HW}{16}\times d}$ represents the seeds after passing through the self-attention layers, where $d$ denotes the dimension of the seeds.

The OVM is aimed at vectorizing semantically independent regions into tokens while achieving arbitrary adjustment of the token quantity. Specifically, we define $N$ learnable vectors $q$ as queries and utilize the cross-attention mechanism to merge seeds belonging to the same semantically independent region into homogeneous tokens. Here, $N$ acts as a variable hyperparameter to enable arbitrary adjustment of the token quantity (as detailed in Section \ref{method_OVM}). This process can be formalized as follows:
\begin{equation}
    t = V(f,q)
\end{equation}
where $q\in \mathbb{R}^{N \times D}$ and where $V$ represents the object vectorization module.

In conclusion, the entire tokenizer is formalized as
\begin{equation}
    t = T(I) = V(P(I),q)
\end{equation}

\subsection{Object Perception Module}
\label{method_OPM}

HOOK treats semantically independent regions as the basic elements of vision. How do we find SIRs in an image? To gain inspiration from a similar problem in NL tokenization, namely, finding semantically meaningful subwords, we examine an elegant solution provided by Byte-Pair Encoding (BPE)\cite{sennrich2015neural}: First, we split the language into the smallest elements, such as by splitting on the basis of letters. Second, we calculate the most frequently occurring letter combinations in the corpus and construct new tokens on the basis of those combinations, iterating this process. This approach corresponds to the "splitting and merging" route mentioned in Section \ref{theoretical_analysis_general_routing}. The OPM consists of three main steps: 1. splitting the image into several fine-grained seeds; 2. merging the seeds into SIRs; and 3. stopping and merging.

\textbf{Step 1: Splitting the image into several fine-grained seeds}

One of the most intuitive approaches inspired by Byte-Pair Encoding (BPE) is to treat each pixel in the image as a seed and then to gradually merge them to form SIRs. However, the cost of pixels as initial seeds is too high, as a 224 $\times$ 224 image can have as many as 50,176 pixels. To address this issue, we introduce a hypothesis specifically for remote sensing images:

\begin{assumption}
\label{assumption1}
    In remote sensing images, there are no complete objects within a 4$\times$4 pixel window.
\end{assumption}

Under Assumption \ref{assumption1}, the size of any SIR is larger than 4$\times$4, so dividing the image into several 4$\times$4 pixel-sized seeds actually converts the scenario from \textit{"multiple tokens multiple objects"} to \textit{"same object multiple tokens"}, as shown in Figure \ref{fig:confusionMatrix} from square 4 to square 2.

In the implementation, we use convolutional modules to obtain the seeds, as described above. Specifically, we stack 2 convolutional layers with a kernel size of 2 and a stride of 2 to extract a local feature vector from the 4 $\times$ 4-pixel window, which becomes the obtained seed. Additionally, to enhance the representation capability of the seeds and support the expansion of subsequent SIRs, we adopt the traditional CNN architecture and intersperse convolutional layers with a kernel size of 3, a stride of 1, and a padding of 1 with batch normalization layers and rectified linear unit (ReLU) activation layers within the convolutional block. At the end of the convolutional block, we add a dimension projection layer with a kernel size of 1 and a stride of 1 to project the image features to a 512-dimensional space.

\textbf{Step 2: Merging the seeds into SIRs}

How can seeds be expanded into SIRs? In natural language processing, a self-attention mechanism essentially relates and aggregates tokens, making tokens with semantic relationships more similar. This phenomenon has also been observed in various visual data\cite{wu2020visual, ru2023token, zhou2022token, gong2021vision}. Inspired by this finding, we attempt to introduce a self-attention layer to make the seeds belonging to the same SIR more similar to achieve expansion from seeds to SIRs.

Considering that the token sequence from step 1 is much longer than that of Patch Embed and that multiple works\cite{Beltagy2020Longformer, liu2021Swin, li2022exploring, kirillov2023segment} have reported that self-attention suffers from decreased efficiency and attention degradation when processing long sequences, we adopt the SAM\cite{kirillov2023segment, rw2019timm} instead of the standard self-attention module.

Specifically, we introduce local attention and global attention into the SAM. First, local attention is utilized to establish relationships among seeds within a local window, which is typically constrained to a quarter of the image. Second, global attention expands these relationships to a global scale, functioning similarly to a standard self-attention layer. The ablation experiments in Section \ref{experiment_ablation} demonstrate that local attention, global attention, and their stacking all play important roles in the overall performance of HOOK.

\textbf{Step 3: Stopping and merging}

According to the analysis in Section \ref{theoretical_analysis_why}, the ideal granularity of semantically independent regions depends on the specific image and task. Therefore, in the stopping strategy for the continuous merging of seeds into SIRs, we did not introduce additional constraints or guidance. The expansion process of seeds naturally stops after passing through one layer of local attention and one layer of global attention, and it is jointly optimized with the entire model under the supervision of a specific task loss function.

\begin{figure*}[!t]
	  \centering
	\includegraphics[width=1.0\textwidth]{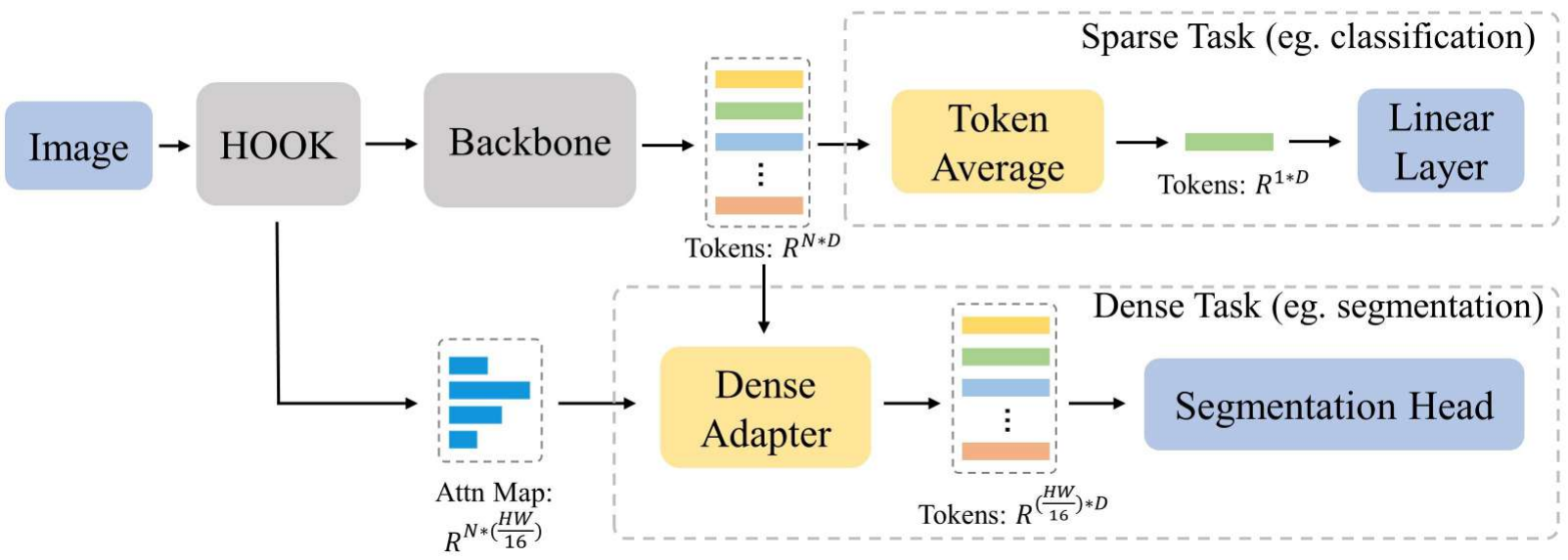}
        \caption{HOOK is capable of adapting to tasks with different granularities. For sparse tasks represented by classification, the model can average the tokens and then pass them through a linear classification layer to output the classification results. For dense tasks represented by segmentation, the model can utilize the intermediate variable, namely, the attention map from the OVM, to restore the number of tokens and then pass them to the segmentation head to output the segmentation results.}
\label{fig:HOOK_task}
\end{figure*}

\subsection{Object Vectorization Module}
\label{method_OVM}

The OPM merges seeds into SIRs, where seeds belonging to the same SIR will be more similar. The OVM is aimed at vectorizing SIRs into visual tokens, and the number of visual tokens can be arbitrarily adjusted to meet adaptability requirements.

A. Jaegle et al. proposed a perception model framework called the Perceiver\cite{jaegle2021perceiver}, which consists of only cross-attention and self-attention layers. This model can perceive multiple modalities without changing its structure. The core idea is that the modality data serve as the keys and values for cross-attention, whereas the model's predefined latent variables serve as queries. The latent variables are repeatedly exposed to the original modality data through stacked cross-attention, continuously refining the model's perceptual results. Moreover, because only the latent variables pass through the self-attention layer, the model's efficiency is greatly improved.

The success of the Perceiver inspires us in two ways: (1) The predefined query in cross-attention has the ability to perceive semantic information in the key and value. (2) The number of tokens obtained through cross-attention is determined only by the query.

Furthermore, from the calculation principle of cross-attention, as indicated by Equation \ref{eq_crossattn}, where $Q$ denotes predefined queries, $K$ and $V$ denote outputs of the OPM, $\sigma$ denotes the Softmax operation, and $d$ denotes the dimension of $Q$, the operation of calculating the similarity between the query and the key matches the way the OPM finds SIRs (i.e., making seeds belonging to the same SIR more similar).

\begin{equation}
\label{eq_crossattn}
    v^\prime = \sigma(\frac{QK^T}{\sqrt d})V
\end{equation}

Therefore, we implement the object vectorization module on the basis of cross-attention. Specifically, we define $N$ learnable vectors $q$ as the queries for cross-attention, while the seeds output by the object perception module serve as the keys and values. During the forward process, each vector in $q$ queries the seeds to retrieve the most similar seed set and aggregates them with weights to form visual tokens.

In the specific implementation, we empirically find that when the dimension of cross-attention is defined as half of the object perception module's dimension, i.e., 256, HOOK performs the best. Additionally, to align the dimensions of the visual tokens with the subsequent backbone network dimensions, we add an MLP after the cross-attention layer. The dimensions of the MLP and the learnable vectors q are set to the dimensions of the backbone network. For example, if the backbone network is a standard ViT, the dimension is 768.

\subsection{HOOK is image-agnostic and task-agnostic}
\label{method_task}

HOOK is independent of specific images and tasks and is both image-agnostic and task-agnostic.

First, HOOK can accept images of any size. HOOK uses a convolutional block to extract local features from an image (as detailed in Section \ref{method_OPM}). Since convolutional operations are not sensitive to image size, HOOK can handle images of any size. Considering that a large-scene image may lead to too many seeds in a 4$\times$4 pixel window, which would increase the computational cost of the two self-attention layers in the OPM, we implement adjustable seed sizes. In the experiments in Section \ref{experiment_main}, we demonstrated the changes in efficiency and accuracy due to variations in the convolution window size. We showed that even for large-scale remote sensing images, HOOK maintains high efficiency and accuracy.

Furthermore, HOOK demonstrates excellent performance on both dense tasks and sparse tasks. In sparse prediction tasks such as classification, the model needs to output only sparse discriminative information for the original image. As shown in Figure \ref{fig:HOOK_task}, after the tokens output by HOOK pass through the backbone network, they can directly pass through a linear classification layer to produce classification results. In dense prediction tasks such as semantic segmentation, the model needs to output not only discriminative information but also pixel-level spatial information. We observe that the cross-attention in the OVM can output not only visual tokens but also intermediate variables, namely, attention maps that indicate the region represented by each visual token (i.e., the region of the token mentioned in Section \ref{theoretical_analysis_confusionMatrix}). Therefore, we use the attention map to reconstruct the original number of visual tokens, thus restoring spatial information. The experiments in Section \ref{experiment_main} demonstrate that HOOK requires only 8 tokens to perform the semantic segmentation task.

\section{Experiments}
\label{experiments}

The experiments demonstrate that HOOK possesses two fundamental properties, namely, homogeneity and adaptability, and exhibits excellent performance and efficiency.

Section \ref{experiment_setup} describes how we organized the experiments, including the settings, datasets, and baselines. Section \ref{experiment_main} and Section \ref{experiment_comparsion_pretrained} present HOOK's results for sparse and dense tasks on five datasets, with quantitative comparisons in terms of accuracy and efficiency with the previous visual tokenizer and pretrained Patch Embed. The visualization results for the homogeneity of the visual tokens are shown in Section \ref{experiment_vis}.

Additionally, we conducted a more in-depth analysis of the phenomena observed in the experiments. Section \ref{experiment_stack} experimentally verifies the simple intuition that under a stronger visual tokenizer, deeper backbone networks are redundant. Section \ref{experiment_deeper_tokenizer} discusses whether deeper visual tokenizers perform better. The ablation results and analysis of the two types of attention mechanisms within the OVM are presented in Section \ref{experiment_ablation}. In Section \ref{experiment_limitation}, we discuss several limitations of HOOK to provide a comprehensive view of the HOOK method.

\subsection{Setup}
\label{experiment_setup}

\begin{table*}[!t]
\caption{Results of HOOK on sparse and dense tasks}
\label{tab:exp_main}
\resizebox{\textwidth}{!}{%
\begin{tabular}{@{}cccccccccccc@{}}
\toprule
\multirow{2}{*}{\textbf{Tokenizer}} &
  \multicolumn{6}{c|}{\textbf{Sparse Task (Classification)}} &
  \multicolumn{5}{c}{\textbf{Dense Task (Segmentation)}} \\ \cmidrule(l){2-12} 
 &
  \textbf{\begin{tabular}[c]{@{}c@{}}Number of\\ Tokens\end{tabular}} &
  \textbf{\begin{tabular}[c]{@{}c@{}}Params\\ (M)\end{tabular}} &
  \textbf{\begin{tabular}[c]{@{}c@{}}MACs\\ (G)\end{tabular}} &
  \textbf{\begin{tabular}[c]{@{}c@{}}NWPU\\ (Acc1)\end{tabular}} &
  \textbf{\begin{tabular}[c]{@{}c@{}}RS19\\ (Acc1)\end{tabular}} &
  \multicolumn{1}{c|}{\textbf{\begin{tabular}[c]{@{}c@{}}NaSC-TG2\\ (Acc1)\end{tabular}}} &
  \textbf{\begin{tabular}[c]{@{}c@{}}Number of\\ Tokens\end{tabular}} &
  \textbf{\begin{tabular}[c]{@{}c@{}}Params\\ (M)\end{tabular}} &
  \textbf{\begin{tabular}[c]{@{}c@{}}MACs\\ (G)\end{tabular}} &
  \textbf{\begin{tabular}[c]{@{}c@{}}GID5\\ (mIoU)\end{tabular}} &
  \textbf{\begin{tabular}[c]{@{}c@{}}DGLCC\\ (mIoU)\end{tabular}} \\ \midrule
Patch Embed &
  196 &
  85.68 &
  16.86 &
  70.30 &
  78.10 &
  \multicolumn{1}{c|}{86.56} &
  1024 &
  121.47 &
  206.94 &
  67.79 &
  54.57 \\ \midrule
Quadtree &
  100 &
  86.02 &
  8.82 &
  70.99 &
  81.05 &
  \multicolumn{1}{c|}{86.55} &
  400 &
  123.28 &
  158.77 &
  49.18 &
  39.62 \\
VT &
  8 &
  96.27 &
  2.59 &
  71.95 &
  80.07 &
  \multicolumn{1}{c|}{84.24} &
  8 &
  133.12 &
  132.83 &
  58.12 &
  45.61 \\
PnP-DETR &
  54 &
  111.97 &
  8.94 &
  73.53 &
  81.37 &
  \multicolumn{1}{c|}{85.91} &
  158 &
  148.97 &
  158.82 &
  62.12 &
  50.04 \\
Conv-VGG19 &
  49 &
  105.51 &
  23.78 &
  74.46 &
  83.66 &
  \multicolumn{1}{c|}{80.60} &
  256 &
  142.62 &
  247.18 &
  53.73 &
  47.44 \\
Conv-ResNet50 &
  49 &
  110.17 &
  8.46 &
  75.17 &
  81.37 &
  \multicolumn{1}{c|}{86.47} &
  256 &
  147.28 &
  167.15 &
  68.81 &
  48.72 \\ \midrule
HOOK-32 &
  6 &
  105.51 &
  5.87 &
  71.73 &
  85.29 &
  \multicolumn{1}{c|}{84.74} &
  8 &
  141.31 &
  147.66 &
  60.98 &
  48.01 \\
HOOK-16 &
  6 &
  99.22 &
  6.80 &
  74.13 &
  84.97 &
  \multicolumn{1}{c|}{86.94} &
  8 &
  135.01 &
  152.70 &
  73.53 &
  54.08 \\
HOOK-8 &
  6 &
  97.65 &
  10.66 &
  77.21 &
  86.60 &
  \multicolumn{1}{c|}{87.26} &
  8 &
  133.44 &
  172.87 &
  75.79 &
  \textbf{54.72} \\
HOOK-4 &
  6 &
  97.25 &
  26.10 &
  \textbf{77.38} &
  \textbf{87.58} &
  \multicolumn{1}{c|}{\textbf{88.25}} &
  8 &
  133.05 &
  253.52 &
  \textbf{78.81} &
  54.27 \\ \bottomrule
\end{tabular}%
}
\end{table*}

A complete transformer-based visual model should consist of three main components: a visual tokenizer, a transformer backbone, and a task head. The visual tokenizer is responsible for converting raw images into token sequences, with the most popular visual tokenizer currently being Patch Embed. The transformer backbone serves as the core of the model, extracting and understanding the semantic information within the token sequences. The task head aligns the output from the backbone with specific downstream tasks. In this paper, we propose a new visual tokenizer called HOOK, which aims to replace Patch Embed to enhance the model's ability to understand visual data.

To fairly and effectively test the performance of HOOK, unless specified otherwise, we use the same transformer backbone (12-layer, 768-dimensional transformer encoder) and task heads (linear classifier for classification tasks and SegFormer\cite{xie2021segformer} head for segmentation tasks) in all the experiments. Notably, many improvements to visual tokenizers not only change the structure of the visual tokenizer but also introduce new modules or mechanisms into the backbone or task head\cite{yin2021adavit, rao2021dynamicvit, chen2023cf, beyer2023flexivit, yuan2021tokens}. To accurately assess the impact of the visual tokenizer on the model's performance, our baseline for comparison excludes these types of methods and only includes works that focus solely on improving the visual tokenizer.

For the credibility of our experimental results, our first principle in selecting datasets is that they must be publicly available and widely researched. Additionally, we aim to validate that HOOK is image-agnostic and task-agnostic while testing its performance. Therefore, we test the performance of HOOK on both sparse tasks, represented by classification, and dense tasks, represented by segmentation. Specifically, for the classification tasks, we choose NWPU-RESISC45\cite{cheng2017remote}, WHU-RS19\cite{Xia2010WHURS19, Dai2011WHURS19}, and NaSC-TG2\cite{NaSCTG2}, with all datasets having a fixed image size of 224$\times$224. For the segmentation tasks, we select GID5\cite{Tong2020GID} and DGLCC\cite{DGLCC}, which have a fixed image size of 512$\times$512.

We use the top-1 accuracy as the evaluation metric for the classification task and the mean intersection over union (mIoU) as the evaluation metric for the semantic segmentation task. We also quantitatively compare the efficiency between different methods, choosing the number of multiply accumulate operations (MACs) as the metric. In all our experiments, the number of MACs refers to the number of multiply accumulate operations during one forwards process of the entire model.

For conciseness, details on dataset preprocessing, experimental hyperparameters, and other specific information are provided in \ref{appendix:exp_details}. Additionally, all executable code, data, and hyperparameters are available in the \href{https://github.com/GeoX-Lab/Hook}{GitHub repository}.

\subsection{Main results}
\label{experiment_main}

We test the performance of HOOK on two categories of tasks: sparse tasks and dense tasks. The essential difference between the two types of tasks is that sparse tasks require only sparse discriminative information, such as categories, whereas dense tasks require not only discriminative information but also dense spatial information. Representative sparse tasks include classification, reidentification, and sentiment analysis, whereas representative dense tasks include segmentation, detection, and depth estimation.

In addition to the 4$\times$4 pixel seed size, to balance accuracy and efficiency, we also test different HOOK versions with seed sizes of 8$\times$8, 16$\times$16, and 32$\times$32 pixels. As shown in Table \ref{tab:exp_main}, these versions are denoted HOOK-4, HOOK-8, HOOK-16, and HOOK-32, respectively.

To accurately evaluate the effects of different visual tokenizers on the overall performance of the model, the baseline for comparison includes only works that optimize the visual tokenizer while keeping the transformer backbone and task head consistent.

Among our selected baseline methods, Quadtree\cite{ronen2023vision} and PnP-DETR\cite{wang2021pnp} are patch-based methods. The former assumes that all patches in the image are unreasonable and uses GradCAM to identify significant regions, implementing the generation of patches with different granularities via quadtrees. The latter reveals that the number of patches is redundant, and a scoring mechanism is used to identify background and foreground regions and merge features of background areas. VT\cite{wu2020visual} is an object-oriented method that uses a simple convolution operation to assign each pixel in the image to one of several semantic groups, which are eventually mapped to a visual token. Additionally, several studies\cite{xiao2021early, wang2021pnp} have suggested that the combination of convolution and a transformer leads to better performance for transformer-based models, so we also tested the performance of two classic convolutional networks (VGG\cite{simonyan2014very} and ResNet\cite{he2016deep}) as visual tokenizers.

\begin{table*}[!t]
\caption{Comparison results of pretrained Patch Embed and HOOK with different ViT architectures}
\label{tab:exp_pretrained}
\resizebox{\textwidth}{!}{%
\begin{tabular}{@{}c|ccccc|cc@{}}
\toprule
\textbf{Visual Tokenizer}      & \textbf{Pretrained Model} & \textbf{Publication} & \textbf{Arch.}                   & \textbf{Params (M)} & \textbf{MACs (G)} & \textbf{NWPU}           & \textbf{RS19}           \\ \midrule
\multicolumn{1}{c|}{\multirow{19}{*}{Patch Embed}} & ViT\cite{dosovitskiy2020image}                & ICLR2023 & \multirow{2}{*}{ViT-Small} & 21.61 & 4.25  & 76.04 & 82.68 \\
\multicolumn{1}{c|}{} & DINO-MC\cite{wanyan2023dinomc}          & Arxiv2023   &                        & 21.39      & 16.73    & 73.61          & 78.76          \\
\cmidrule(l){2-8} 
\multicolumn{1}{c|}{} & ViT\cite{dosovitskiy2020image}              & ICLR2021    & \multirow{7}{*}{ViT-Base} & 52.68      & 16.86    & 77.21          & 84.64          \\
\multicolumn{1}{c|}{} & BEiT\cite{beit}             & ICLR2022    &                        & 85.68      & 16.86    & 73.75          & 82.35          \\
\multicolumn{1}{c|}{} & CLIP\cite{radford2021learning}             & PMLR2021    &                        & 85.68      & 16.86    & 74.06          & 83.33          \\
\multicolumn{1}{c|}{} & DeiT\cite{pmlr-v139-touvron21a}             & ICML2021    &                        & 85.68      & 16.86    & 74.89          & 83.66          \\
\multicolumn{1}{c|}{} & MTP\cite{MTP}              & JSTARS2024  &                        & 85.68      & 16.78    & 74.27          & 83.01          \\
\multicolumn{1}{c|}{} & CMID\cite{muhtar2023cmid}             & TGRS2023    &                        & 85.68      & 16.78    & 72.40          & 79.08          \\
\multicolumn{1}{c|}{} & CSPT\cite{zhang2022consecutive}             & RS2022      &                        & 85.68      & 16.78    & 71.82          & 79.74          \\ \cmidrule(l){2-8} 
\multicolumn{1}{c|}{} & ViT\cite{dosovitskiy2020image}              & ICLR2021    & \multirow{7}{*}{ViT-Large} & 303.14     & 59.68    & 74.94          & 84.31          \\
\multicolumn{1}{c|}{} & BEiT\cite{beit}             & ICLR2022    &                        & 303.14     & 59.68    & 72.98          & 84.97          \\
\multicolumn{1}{c|}{} & CLIP\cite{radford2021learning}             & PMLR2021    &                        & 302.96     & 77.82    & 73.28          & 84.31          \\
\multicolumn{1}{c|}{} & SatMAE\cite{satmae2022}           & NIPS2022    &                        & 303.14     & 59.38    & 74.23          & 83.66          \\
\multicolumn{1}{c|}{} & SatMAE-t\cite{satmae2022}         & NIPS2022    &                        & 303.14     & 59.38    & 72.25          & 79.74          \\
\multicolumn{1}{c|}{} & MTP\cite{MTP}              & JSTARS2024  &                        & 303.14     & 59.38    & 73.46          & 84.97          \\
\multicolumn{1}{c|}{} & ScaleMAE\cite{reed2023scale}         & ICCV2023    &                        & 303.14     & 59.38    & 73.16          & 82.68          \\ \cmidrule(l){2-8} 
\multicolumn{1}{c|}{} & ViT\cite{dosovitskiy2020image}              & ICLR2021    & \multirow{3}{*}{ViT-Huge} & 630.49     & 161.97   & 75.86          & 87.58          \\
\multicolumn{1}{c|}{} & CLIP\cite{radford2021learning}             & PMLR2021    &                        & 630.49     & 161.97   & 73.65          & 84.31          \\
\multicolumn{1}{c|}{} & SAM\cite{kirillov2023segment}              & ICCV2023    &                        & 630.72     & 123.57   & 73.13          & 83.66          \\ \midrule
\multicolumn{1}{c|}{\multirow{4}{*}{HOOK}}         & \multirow{4}{*}{-} & \multirow{4}{*}{-} & ViT-Small                  & 29.74 & 25.69 & 77.32 & 87.91 \\
\multicolumn{1}{c|}{} &                  &             & ViT-Base                  & 97.25      & 26.10    & \textbf{77.38} & 87.58          \\
\multicolumn{1}{c|}{} &                  &             & ViT-Large                  & 318.32     & 27.42    & 77.17          & \textbf{87.91} \\
\multicolumn{1}{c|}{} &                  &             & ViT-Huge                  & 650.55     & 29.41    & 76.73          & 86.27          \\ \bottomrule
\end{tabular}%
}
\end{table*}

\textbf{A. Sparse Tasks}

For sparse tasks, we choose classification on the NWPU-RESISC45, WHU-RS19, and NaSC-TG2 datasets; as shown in Table \ref{tab:exp_main}, HOOK requires only 6 tokens to outperform the standard Patch Embed by margins of 7.08\%, 9.48\%, and 1.69\%, respectively, on these datasets. Compared with the baseline models, HOOK achieves state-of-the-art performance in terms of both the number of tokens used and the classification accuracy.

\textbf{B. Dense Tasks}

For dense tasks, we focus on semantic segmentation on the GID5 and DGLCC datasets. The dense adapter module, introduced in Section \ref{method_task}, addresses the challenge of completing dense tasks with a small number of tokens without requiring additional training parameters. As shown in Table \ref{tab:exp_main}, HOOK requires only 8 tokens to outperform the standard Patch Embed, with margins of 11.02\% and 0.15\% on the GID5 and DGLCC datasets, respectively. Compared with the baseline models, HOOK achieves state-of-the-art performance in terms of both the number of tokens used and the segmentation accuracy.

\textbf{C. Efficiency}

In addition to accuracy, the efficiency of the visual tokenizer is crucial. As shown in Table \ref{tab:exp_main}, we compare the numbers of parameters and MACs between HOOK and the baseline methods.

Although the structure of HOOK is more complex than that of Patch Embed, HOOK significantly reduces the number of tokens, which greatly improves the efficiency of the backbone network. The results in Table \ref{tab:exp_main} indicate that as the pixel window size increases, the model's efficiency improves, whereas the accuracy gradually decreases. In the classification task, HOOK-8 demonstrates a distinct advantage in efficiency over Patch Embed in terms of the MACs metric (10.66 vs. 16.86), with only a slight decrease in accuracy compared with HOOK-4 (0.17 on NWPU) and still outperforming Patch Embed (6.91). This trend can be observed across other datasets and suggests that HOOK has distinct advantages over the baseline methods in terms of both accuracy and efficiency.

\subsection{Comparison with pretrained Patch Embed}
\label{experiment_comparsion_pretrained}

State-of-the-art visual models are typically pretrained on large-scale datasets. Examples include ViT\cite{dosovitskiy2020image} and MTP\cite{MTP}, which are supervised pretrained on labelled data, as well as DINO-MC\cite{wanyan2023dinomc} and SatMAE\cite{satmae2022}, which are self-supervised pretrained on unlabelled data. In Table \ref{tab:exp_pretrained}, we compare the performance of randomly initialized HOOK and pretrained Patch Embed on the NWPU-RESISC45 and WHU-RS19 datasets. Note that we only load the weights of Patch Embed from the pretrained models to fairly compare the effects of different visual tokenizers, while the backbones are randomly initialized. Following the configurations of the original pretrained models, we compare four backbone architectures at different scales: ViT-Small, ViT-Base, ViT-Large, and ViT-Huge. The main differences among these architectures are shown in Table \ref{tab:exp_arch}.

\begin{table}[htbp]
\centering
\caption{Comparison of different ViT architectures}
\label{tab:exp_arch}
\begin{tabular}{cccc}
\hline
\textbf{Arch.} & \textbf{Layers} & \textbf{Dimension} & \textbf{Heads} \\ \hline
ViT-Small         & 12              & 384                & 4              \\
ViT-Base          & 12              & 768                & 8              \\
ViT-Large         & 24              & 1024               & 8              \\
ViT-Huge          & 32              & 1280               & 10             \\ \hline
\end{tabular}
\end{table}

Table \ref{tab:exp_pretrained} demonstrates that the accuracy of the randomly initialized HOOK is superior to that of the pretrained Patch Embed across various backbone architectures. Furthermore, owing to HOOK's significant reduction in the number of visual tokens, its efficiency advantage becomes more pronounced as the backbone size increases. For example, when the ViT-Huge backbone is used, the model's number of MACs is only 29.41, which is significantly lower than the standard ViT-Huge's 161.97. In the context of large multimodal models, this characteristic can greatly reduce training and inference costs.

\subsection{Visualization of homogeneous tokens}
\label{experiment_vis}

\begin{figure*}[htbp]
	  \centering
	\includegraphics[width=1.0\textwidth]{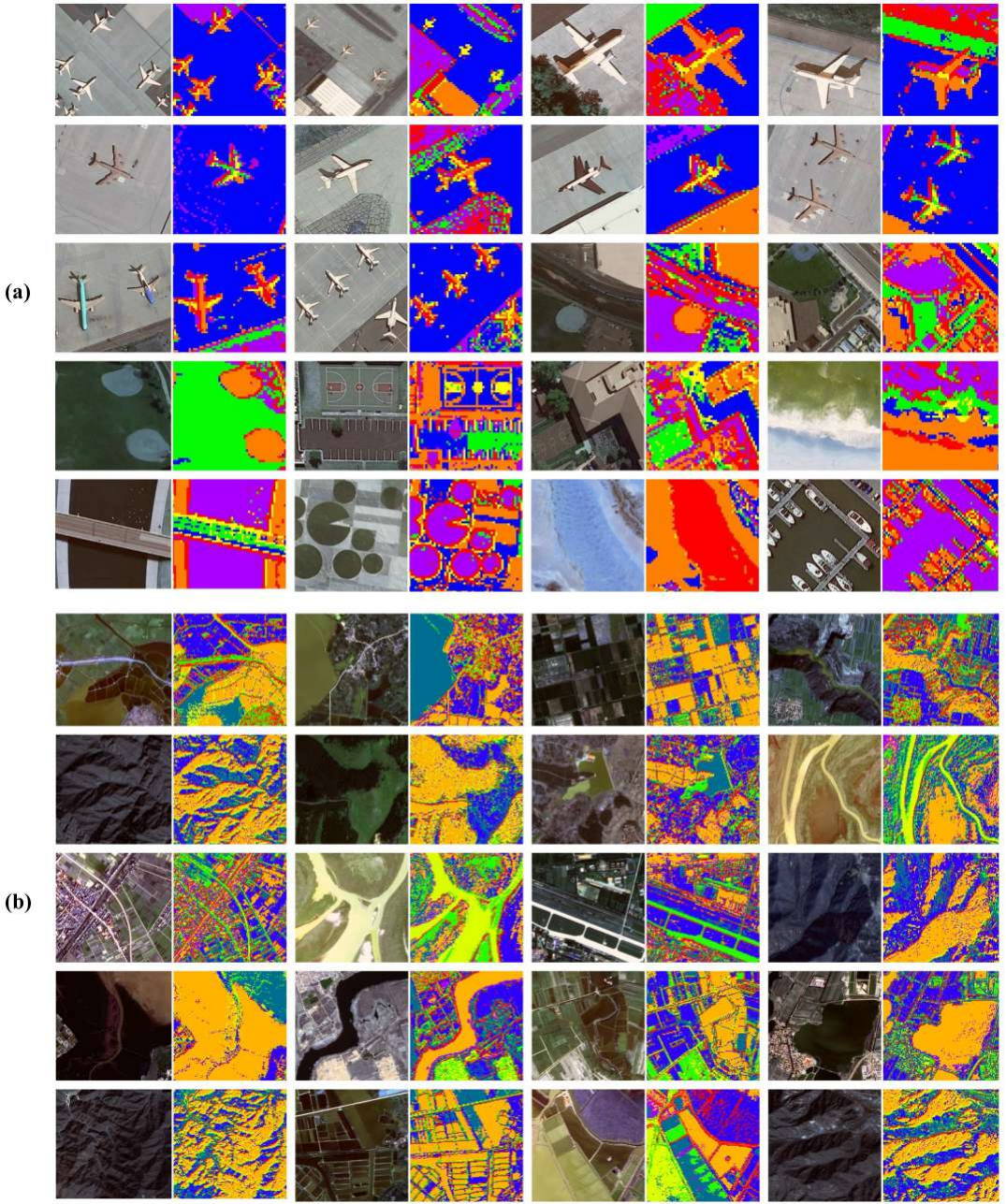}
        \caption{Visualization results for homogeneous visual tokens. The left image is the original image, whereas the right image displays the regions of the token, with each colour representing one region of the token. The images in (a) are obtained from the NWPU-RESISC45 classification dataset, whereas the images in (b) are obtained from the GID5 semantic segmentation dataset.}
\label{fig:vis}
\end{figure*}

\begin{figure*}[htbp]
	  \centering
	\includegraphics[width=0.95\textwidth]{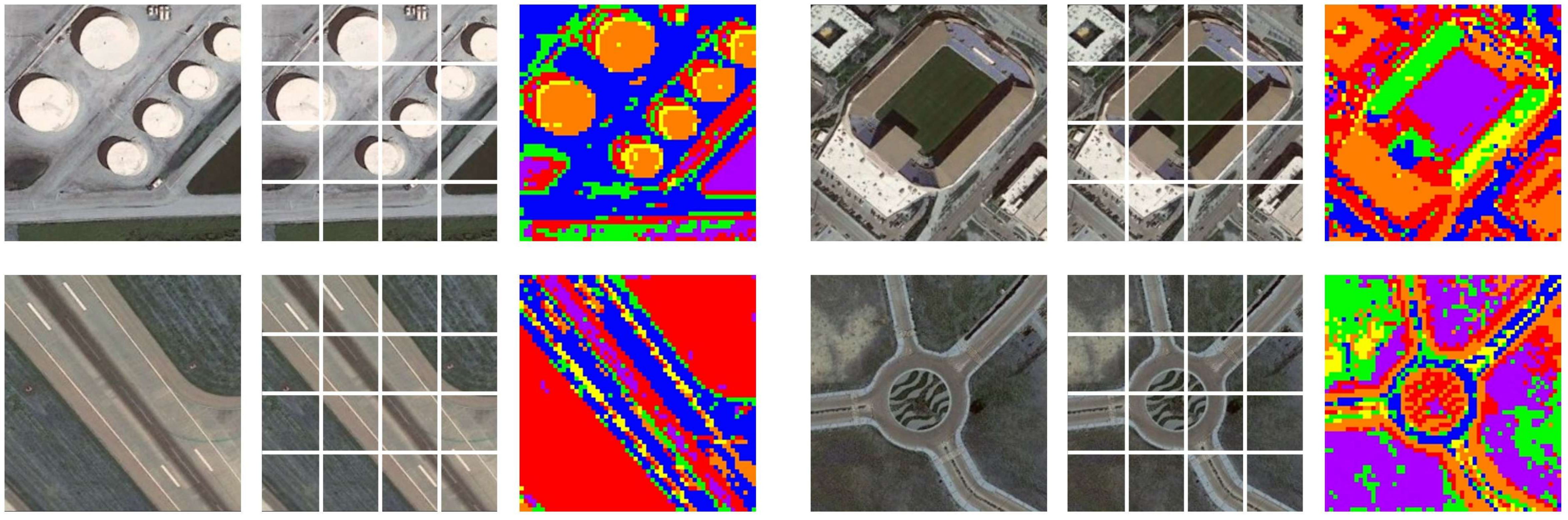}
        \caption{Visualization comparison between regions of the token in HOOK and those in the patch-based tokenizers. Visual tokens in HOOK can adaptively adjust according to the shapes of the objects rather than being regular rectangular patches.}
\label{fig:vis_comp}
\end{figure*}

In Section \ref{method_OVM}, we mentioned that HOOK introduces a cross-attention mechanism that utilizes $N$ learnable query vectors to aggregate image features into $N$ tokens. We save the intermediate variable, namely, the attention map, which records the correspondence between the $N$ tokens output by HOOK and the original image features. By visualizing the attention map, we can observe the regions of the tokens.

Figure \ref{fig:vis} displays the visualization results, where each colour represents one region of each token. The results indicate that each token corresponds to a specific object and that the regions of the token can adaptively adjust according to the shape of the object rather than being a regular rectangular patch. Figure \ref{fig:vis_comp} shows a comparison between regions of the token in HOOK and those in Patch Embed, which demonstrates that HOOK can describe the information of objects in the image more clearly with fewer tokens. For more visualization results, please refer to \ref{appendix:Supplementary_vis}.

\subsection{Stacking up to 12 layers is redundant}
\label{experiment_stack}

The standard ViT model consists of two main modules, Patch Embed and 12 layers of self-attention modules, with the primary computational cost concentrated in the latter. Intuitively, if a stronger visual tokenizer replaces Patch Embed, the 12 layers of self-attention modules may become redundant.

To validate this idea, we conduct tests on classification tasks, and Table \ref{tab:exp_stack_cls} displays the experimental results. In Table \ref{tab:exp_stack_cls}, boldface indicates the best performance, and underlining indicates the second-best performance. The experimental results show that when the number of backbone layers is reduced from 12 layers to 1 layer, HOOK results in less accuracy loss than models that use Patch Embed as the visual tokenizer. Specifically, on the NWPU dataset, Patch Embed incurs a loss of 3.85, whereas HOOK incurs a loss of only 1.93. On the WHU-RS19 dataset, Patch Embed incurs a loss of 3.26, whereas HOOK incurs a loss of only 0.65. These experimental results preliminarily confirm our intuition.

\begin{table}[htbp]
\centering
\caption{Classification with fewer backbone layers}
\label{tab:exp_stack_cls}
\begin{tabular}{@{}lcccc@{}}
\toprule
\textbf{Tokenizer} & \textbf{Layers} & \textbf{Params(M)} & \textbf{NWPU} & \textbf{RS19} \\ \midrule
\multirow{3}{*}{Patch Embed} & 12 & 85.68         & 70.30          & 78.10          \\
                             & 1  & \textbf{7.71} & 66.45          & 74.84          \\
                             & 3  & 21.89         & 69.72          & 78.76          \\ \midrule
\multirow{2}{*}{HOOK(ours)}  & 1  & {\ul 20.90}   & {\ul 76.88}    & {\ul 86.93}    \\
                             & 12 & 98.86         & \textbf{78.81} & \textbf{87.58} \\ \bottomrule
\end{tabular}
\end{table}

\begin{table}[htbp]
\centering
\caption{Segmentation with fewer backbone layers}
\label{tab:exp_stack_seg}
\begin{tabular}{@{}lccc@{}}
\toprule
\textbf{Tokenizer} &
  \textbf{\begin{tabular}[c]{@{}c@{}} Layers\end{tabular}} &
  \textbf{Params(M)} &
  \textbf{\begin{tabular}[c]{@{}c@{}}GID5 \end{tabular}} \\ \midrule
\multirow{2}{*}{Patch Embed} & 4  & \textbf{64.77} & 69.28          \\
                             & 12 & 121.47         & 67.79          \\ \midrule
\multirow{2}{*}{HOOK(ours)}  & 4  & {\ul 84.74}    & \textbf{78.91} \\
                             & 12 & 141.44         & {\ul 78.81}    \\ \bottomrule
\end{tabular}
\end{table}

HOOK has more parameters than Patch Embed does. To further illustrate that the effectiveness of HOOK stems from its homogeneity and adaptability rather than solely from the increase in the number of parameters, we also compare "Patch Embed + 3 backbone layers" and "HOOK + 1 backbone layer" in Table \ref{tab:exp_stack_cls}. The former model has an overall parameter count of 21.89 M, whereas the latter model has 20.90 M, making their overall numbers of parameters similar. The results in Table \ref{tab:exp_stack_cls} indicate that HOOK outperforms Patch Embed by 7.16 on the NWPU-RESISC45 dataset and by 8.71 on the WHU-RS19 dataset, demonstrating that \textbf{the effectiveness of HOOK does not depend solely on the increase in the number of parameters}.

Interestingly, when Patch Embed is deployed as the visual tokenizer and the number of backbone layers is reduced from 12 layers to 3 layers, the model's accuracy only slightly fluctuates, and in some cases, such as on the WHU-RS19 dataset, it even improves. This phenomenon is even more pronounced in the results for the segmentation task in Table \ref{tab:exp_stack_seg}. Both Patch Embed and HOOK show increases in accuracy, with Patch Embed showing a more significant improvement, further highlighting the redundancy of the 12-layer backbone network. This conclusion may serve as inspiration for the design of future visual backbone models.

\subsection{The deeper the tokenizer is, the better?}
\label{experiment_deeper_tokenizer}

\begin{table*}[!t]
\caption{Results of HOOK with different numbers of OPM and OVM layers}
\label{tab:exp_more_layers}
\resizebox{\textwidth}{!}{%
\begin{tabular}{@{}ccccccc@{}}
\toprule
\textbf{Arch. of OPM} & \textbf{Layers of OPM} & \textbf{Layers of OVM} & \textbf{Params (M)} & \textbf{MACs (G)} & \textbf{Acc.} & $\Delta$ \\ \midrule
L+G     & 2 & 1 & 97.23  & 26.10 & 77.38 & -     \\ \midrule
L$\times$2+G$\times$2 & 4 & 1 & 103.54 & 45.85 & 78.40 & +1.02 \\
(L+G)$\times$2 & 4 & 1 & 103.54 & 45.85 & 78.41 & +1.03 \\
L$\times$3+G$\times$3 & 6 & 1 & 109.84 & 65.61 & 78.25 & +0.87 \\
(L+G)$\times$3 & 6 & 1 & 109.84 & 65.61 & 77.60 & +0.22 \\
L$\times$4+G$\times$4 & 8 & 1 & 116.15 & 85.36 & 77.91 & +0.53 \\
(L+G)$\times$4 & 8 & 1 & 116.15 & 85.36 & 77.82 & +0.44 \\ \midrule
L$\times$3+G$\times$3 & 6 & 1 & 109.84 & 65.61 & 78.25 & +0.87 \\
L$\times$3+G$\times$3 & 6 & 2 & 122.32 & 66.51 & 77.71 & +0.33 \\
L$\times$3+G$\times$3 & 6 & 3 & 134.78 & 67.41 & 77.30 & -0.08 \\
L$\times$4+G$\times$4 & 8 & 1 & 116.15 & 85.36 & 77.91 & +0.53 \\
L$\times$4+G$\times$4 & 8 & 2 & 128.63 & 86.26 & 78.21 & +0.83 \\
L$\times$4+G$\times$4 & 8 & 3 & 141.08 & 87.17 & 77.40 & +0.02 \\ \bottomrule
\end{tabular}%
}
\end{table*}

Following Section \ref{experiment_stack}, a natural question arises: "Do deeper tokenizers perform better?" In Table \ref{tab:exp_more_layers}, we compare the accuracy of HOOK with different depths on the NWPU dataset.

For the OPM, we test two methods of increasing depth. For example, L$\times$3+G$\times$3 represents stacking three layers of local attention followed by three layers of global attention, whereas (L+G)$\times$3 represents alternating local attention and global attention three times. For the OVM, we keep the original Perceiver\cite{jaegle2021perceiver} architecture, stacking self-attention and cross-attention on top of cross-attention. For example, a three-layer OVM consists of one layer of cross-attention followed by two layers of alternating self-attention and cross-attention.

The experimental results in Table \ref{tab:exp_more_layers} indicate that while the deeper HOOK shows slightly better accuracy than the standard version does, the improvement diminishes as the number of layers increases. This suggests that the standard version of HOOK is already the optimal configuration for balancing accuracy and efficiency on the current dataset.

\begin{table*}[htbp]
\centering
\caption{Local and Global Attention}
\label{tab:exp_LG}
\begin{tabular}{lcccc}
\hline
\multirow{2}{*}{\textbf{Tokenizer}} & \multicolumn{2}{c}{\textbf{Object Perception Module}} & \multirow{2}{*}{\textbf{Params (M)}} & \multirow{2}{*}{\textbf{Acc.}} \\ \cline{2-3}
                            & \textbf{Local Attention}  & \textbf{Global Attention} &       &       \\ \hline
\multirow{4}{*}{HOOK} & \XSolidBrush             & \XSolidBrush             & 92.56 & 70.15 \\
                            & \Checkmark & \XSolidBrush             & 95.71 & 75.13 \\
                            & \XSolidBrush            & \Checkmark & 95.71 & 73.74 \\
                            & \Checkmark & \Checkmark & 98.86 & 77.38 \\ \hline
\end{tabular}
\end{table*}

\begin{figure*}[!t]
	  \centering
	\includegraphics[width=0.95\textwidth]{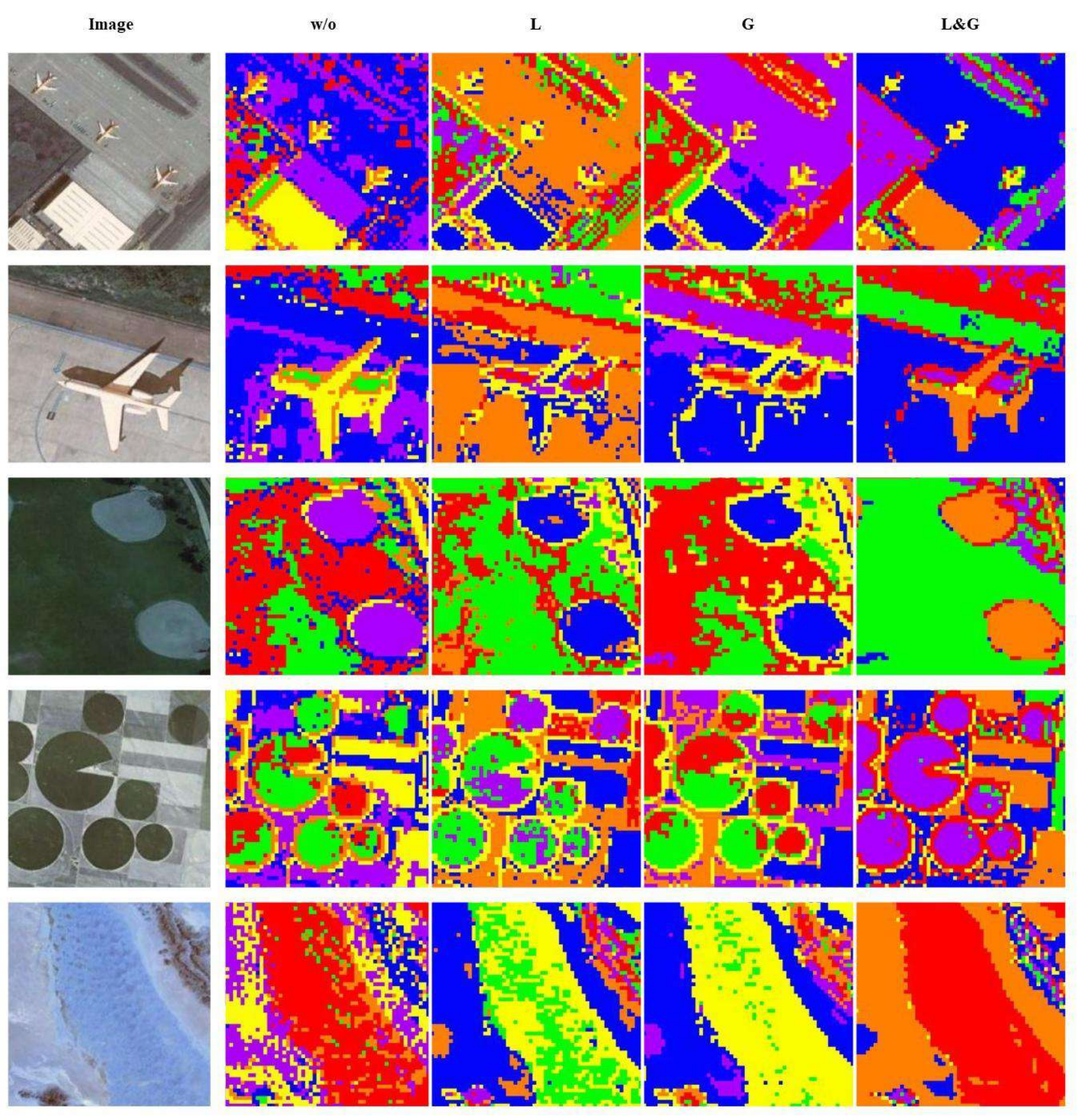}
        \caption{Visualization of the ablation of local and global attention. The homogeneity of visual tokens is optimal when both attention modules are simultaneously utilized.}
\label{fig:LG}
\end{figure*}

\subsection{Ablation}
\label{experiment_ablation}

When designing the OVM, we refer to the implementation of the SAM model and incorporate its local attention module. In Table \ref{tab:exp_LG}, we present ablation experiments conducted on the local attention and global attention modules. In Figure \ref{fig:LG}, we visualize the homogeneous tokens under four different conditions: "no attention module", "local attention only", "global attention only", and "local \& global attention". When either local attention or global attention is lacking, the homogeneity of the visual tokens significantly decreases, leading to a decrease in model accuracy. However, when local attention and global attention are stacked, HOOK is better able to identify SIRs, and the model's accuracy reaches its highest point. This finding demonstrates that local and global attention mechanisms play important roles in improving the homogeneity of visual tokens, thereby influencing the performance of HOOK.

\subsection{Limitations of HOOK}
\label{experiment_limitation}

\begin{figure*}[!t]
	  \centering
	\includegraphics[width=1.0\textwidth]{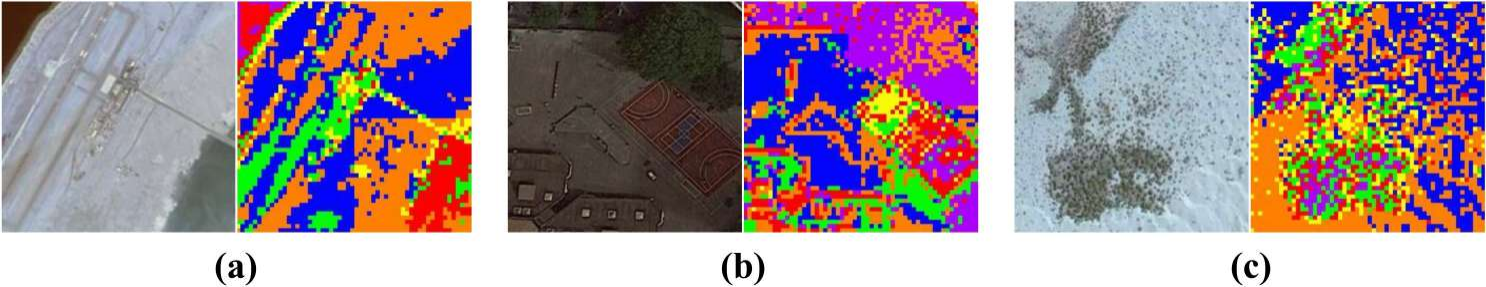}
        \caption{Three typical cases of poor homogeneity: (a) shows that HOOK is limited by the attention range of the local attention module, resulting in poor-homogeneity areas appearing in the lower-right corner; (b) shows that HOOK is influenced by complex colour information and fails to find semantic information in the image; and (c) shows that HOOK is influenced by complex texture information.}
\label{fig:bad_vis}
\end{figure*}

Previous experiments have shown that HOOK has the two properties mentioned above. However, it is important to acknowledge the following limitations of HOOK:

(1) \textbf{Homogeneity}: Ideally, a homogeneous visual tokenizer should rely on semantic information in the image when identifying SIRs. However, from a visualization perspective, HOOK does not completely eliminate the influence of local textures or colours and still considers colour and texture similarity as one of the criteria for determining SIRs. This feature results in poor visualization performance of HOOK on some complex-textured remote sensing images. Additionally, the limitation imposed by local attention also contributes to the decrease in homogeneity. Figure \ref{fig:bad_vis} illustrates three typical scenarios with relatively poor homogeneity: (a) constrained by local attention, (b) overly complex colours, and (c) overly complex textures.

In addition, we do not add any additional guidance or constraints to HOOK. HOOK is only supervised by the downstream task loss function when searching for SIRs. Therefore, we are currently not clear on how adding additional prior assumptions (such as connectivity assumptions) affects the SIRs and the accuracy of the final model. This is one of the directions we will delve into in the future.

\begin{table}[htbp]
    \centering
    \caption{Number of tokens in HOOK}
    \label{tab:exp_tokenNum}
    \begin{tabular}{ccc}
    \hline
    \textbf{Tokenizer} & \textbf{\begin{tabular}[c]{@{}c@{}}Num of Tokens\end{tabular}} & \textbf{Acc.} \\ \hline
    \multirow{5}{*}{HOOK} & 6  & \textbf{77.38} \\
                                & 8  & 77.03          \\
                                & 16 & 77.15          \\
                                & 32 & 76.60          \\
                                & 64 & 76.25          \\ \hline
    \end{tabular}
\end{table}

(2) \textbf{Adaptability}: HOOK can adjust the number of visual tokens as needed. Experiments have shown that it exhibits good adaptability in tasks involving different granularities and images of different sizes. In theory, the larger the number of tokens is, the finer the original image information that is reflected. However, the empirical experiments in Table \ref{tab:exp_tokenNum} show that when we increase the number of tokens in classification tasks, the accuracy of HOOK decreases. This phenomenon indicates that the number of visual tokens may have a more complex nonlinear relationship with the original image and the specific task. This is also an issue that we will delve into in the future.

\section{Discussion}
\label{discussion}

\subsection{The visual tokenizer is the pupil of the machine}
\label{discussion_pupil}

We believe that visual tokenizers play an extremely important role in transformer-based visual models and even in multimodal models. If the visual model is the eye of a machine, then the visual tokenizer is the machine's pupil. Its importance is manifested mainly in the following two aspects:

(1) The visual tokenizer acts as a bridge between the original image and the model. On the one hand, the visual tokenizer needs to directly perceive the high-dimensional original image and compress it into a low-dimensional space to improve model efficiency. On the other hand, the completeness of the perception of the original image information by the visual tokenizer directly determines the upper limit of the model's understanding of the image. In essence, the visual tokenizer compresses the original image into a lossless or low-loss format that conforms to the model's input format (i.e., a token sequence) while striking a balance between efficiency and information completeness.

(2) The visual tokenizer directly affects the granularity and efficiency of the model in understanding images. From the perspective of the original image, visual tokens represent the basic elements of the image. From the model perspective, a unimodal visual model uses the meaning of tokens and the relationships among tokens to understand the image, whereas multimodal models establish relationships among modalities at the token level. Therefore, the quality of the visual tokenizer directly affects the granularity and efficiency of image understanding in visual models and multimodal models.

\subsection{Rethinking HOOK}
\label{discussion_rethink}

HOOK corresponds to the first route in Section \ref{theoretical_analysis_general_routing}: "splitting and merging". First, under the assumption that \textit{"there are no complete objects in a 4$\times$4 pixel window,"} HOOK undergoes extreme splitting of the image. When split to a fine enough level, visual tokens will inevitably satisfy the \textit{"same object multiple tokens"} scenario. We subsequently use a self-attention mechanism to associate seeds belonging to the same SIR and use cross-attention to merge tokens within the same SIR, ultimately achieving the \textit{"same object same token"} scenario.

From this perspective, the main drawback of HOOK is the lack of precision in its splitting operation. In theory, only tokens that overlap with multiple objects need to be split. However, HOOK's rough splitting operation may split tokens that already satisfy homogeneity, making it more difficult to judge whether tokens belong to the same object during subsequent merging operations. This disadvantage reduces the efficiency of the model. The experiment in Table \ref{tab:exp_main} validates this point: in the case where the pixel window size is 4$\times$4, the efficiency of HOOK is lower than that of Patch Embed.

\subsection{Future work}
\label{discussion_future}

Compared to the "splitting and merging" method represented by HOOK, the "merging and splitting" method is equally interesting. One possible implementation is to obtain tokens that can represent all objects in an image and then gradually peel off tokens on the basis of the independence between objects. During this process, we may encounter an entanglement issue between semantically similar objects. Resolving this entanglement is a key challenge that needs to be addressed in this approach.

Furthermore, comparing the two alternative routes, namely, "splitting and merging" and "merging and splitting", is another worthwhile direction. For example, both routes encounter the issue of semantic granularity in the second step. The "splitting and merging" method needs to consider whether tokens that represent the fuselage and wings of an aircraft should be merged into one token during the merging process. On the other hand, the "merging and splitting" method considers whether the token that represents the aircraft should be split into the aircraft and wings. The opposite implementations of the same problem in these two methods may cause fundamental differences in certain application scenarios.

Finally, although we only considered the case of \textit{"same object same token"} as the ideal visual token in the previous analysis, are the scenarios of \textit{"same object multiple tokens"} and \textit{"same token multiple objects"} truly meaningless? For tasks such as human--object interaction detection\cite{kim2021hotr, bergstrom2020human}, which are aimed at gaining a deeper understanding of interaction relationships, could using more tokens to describe the edges of objects in the case of \textit{"same object multiple tokens"} be more suitable? We will further explore these interesting questions in future research.

\section{Conclusions}
\label{conclusion}

Multimodal large-scale language models have led to a revolutionary paradigm shift in the field of remote sensing image understanding. Visual tokenizers, as important and fundamental components, have long been overlooked or even misunderstood. Starting from the essence of these tokenizers, we propose two fundamental properties that an ideal visual tokenizer should possess: (1) homogeneity: semantically independent regions (SIRs) are the basic elements of vision and (2) adaptability: the number of tokens can be arbitrarily adjusted to support images of any size and tasks of any granularity. To construct a visual tokenizer that satisfies the above two properties, we rigorously define and analyse the binary relationship between tokens and objects and derive two general paths for obtaining homogeneous tokens: "splitting and merging" and "merging and splitting". On the basis of the former, we propose a simple HOmogeneous visual tOKenizer, HOOK. The experimental results show that HOOK achieves the \textit{"same token same object"} scenario and outperforms Patch Embed and other baseline methods in both sparse prediction tasks represented by classification and dense prediction tasks represented by segmentation.

This paper emphasizes the importance of visual tokenizers in visual foundation models and multimodal large language models and provides a preliminary theoretical basis for the construction of visual tokenizers. However, the current theoretical analysis and experimental research are still insufficient. For example, the homogeneity that we propose essentially describes the properties of individual tokens, but we have not discussed in detail the emergent behaviour exhibited by combinations of multiple tokens. We hope that our initial work inspires researchers in the field to give attention to and participate in an in-depth investigation of visual tokens.

\bibliographystyle{elsarticle-num-names}
\bibliography{ref}

\onecolumn
\appendix
\section{Supplementary visualization results}
\label{appendix:Supplementary_vis}

\begin{figure*}[!b]
	  \centering
	\includegraphics[width=0.85\textwidth]{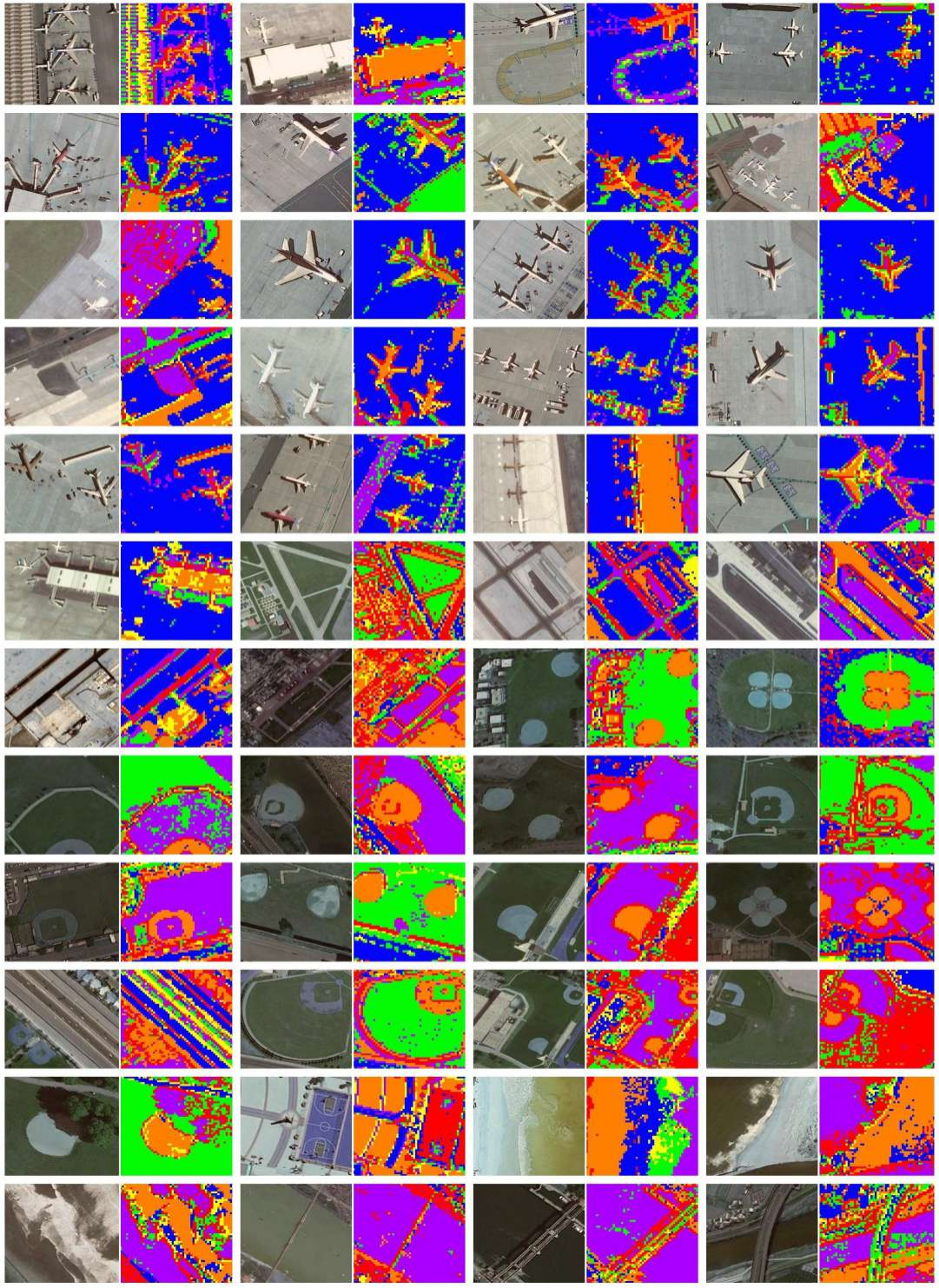}
        \caption{Supplementary visualization results for homogeneous visual tokens}
\label{fig:vis_appendix}
\end{figure*}

Additional visualization results are shown in Figure \ref{fig:vis_appendix}.

\section{Experimental details}
\label{appendix:exp_details}

\subsection{Hyperparameters}
\label{appendix:exp_details_hyperp}

In our experiments, the hyperparameters for training HOOK are kept consistent with those of the baseline methods to ensure a fair comparison. In the classification task, the learning rate is set to 1e-4, the number of training epochs is 100, the learning rate schedule follows a cosine annealing strategy with a 10-epoch warmup, the optimizer is AdamW, the weight decay is 0.05, and the batch size is 32. For the segmentation task, we adjust only the learning rate to 5e-6 and the batch size to 8 while keeping all other hyperparameters the same as those in the classification task.

\subsection{Datasets}
\label{appendix:exp_details_datasets}

For the classification datasets, we randomly select a certain proportion of images from the dataset as the training set, with the remaining images used as the test set. Specifically, the training set ratio is 20\% for the NWPU-RESISC45 dataset, 50\% for the WHU-RS19 dataset, and 5\% for the NaSC-TG2 dataset. For the segmentation datasets, we directly use the dataset splits from TOV\cite{TOV}. Both the GID5 and DGLCC datasets are divided into a training set containing 2000 images of size 512$\times$512 and a test set containing 2500 images of the same size. All the datasets we use can be found at https://github.com/GeoX-Lab/Hook.

\subsection{Metrics}
\label{appendix:exp_details_metrics}

The calculation formula for the top-1 accuracy in the classification task is as follows:

\begin{equation}
\label{eq_top1}
    \text{Top-1 Accuracy} = \frac{N_{\text{correct}}}{N_{\text{total}}} \times 100\%
\end{equation}

where \(N_{\text{correct}} \) represents the number of correct predictions and where \(N_{\text{total}} \) represents the total number of predictions.

The calculation formula for the mean intersection over union (mIoU) in the segmentation task is as follows:

\begin{equation}
\label{eq_mIoU}
    \text{mIoU} = \frac{1}{n} \sum_{i=1}^{n} \frac{A_i^{\text{overlap}}}{A_i^{\text{union}}}
\end{equation}

where $A_i^{\text{overlap}}$ represents the area of overlap between the predicted segmentation and the ground truth for the $i$-th class and where $A_i^{\text{union}}$ represents the area of union of the predicted segmentation and the ground truth for the $i$-th class.

The MACs (multiply-accumulate operations) metric, which is used to measure efficiency, is implemented via the Python library \textit{thop}\cite{thop}. The pseudocode is as follows:

\begin{algorithm}
    \caption{MACs: PyTorch-Like Pseudocode}
    \label{MACs}
    \begin{algorithmic}[1]
        \STATE \texttt{from thop import profile}
        \vspace{10pt}
        \STATE \texttt{model = ViT(tokenizer=HOOK())}
        \STATE \texttt{input = torch.randn((1, 3, 224, 224))}
        \vspace{10pt}
        \STATE \texttt{macs, params = profile(model, inputs=(input,))}
        \STATE \texttt{print(f"MACs: \{macs\}, Params: \{params\}")}
    \end{algorithmic}
\end{algorithm}

\end{document}